\documentclass[11pt]{article}

\usepackage[]{acl}
\usepackage{times}
\usepackage{latexsym}
\usepackage[T1]{fontenc}
\usepackage[utf8]{inputenc}
\usepackage{microtype}
\usepackage{inconsolata}
\usepackage{graphicx}
\usepackage{arydshln}
\usepackage{lscape}
\usepackage{subcaption}
\usepackage{enumitem} 
\usepackage{rotating} 
\usepackage{amsmath}
\usepackage{amssymb}
\usepackage{multirow}
\usepackage{xcolor,pifont}
\usepackage{booktabs}
\usepackage{float}
\usepackage{arydshln}
\usepackage{algorithm}
\usepackage{algpseudocode}
\usepackage{amsmath}
\usepackage{CJKutf8}

\newcommand{\jp}[1]{\begin{CJK}{UTF8}{min}#1\end{CJK}}
\newcommand{\zh}[1]{\begin{CJK}{UTF8}{gbsn}#1\end{CJK}}

\newcommand*\colourcross[1]{%
  \expandafter\newcommand\csname #1cross\endcsname{\textcolor{#1!75}{\ding{56}}}%
}
\colourcross{red}
\newcommand*\colourcheck[1]{%
  \expandafter\newcommand\csname #1check\endcsname{\textcolor{#1!75}{\ding{52}}}%
}
\colourcheck{green}

\newcommand{\ve}[1]{\boldsymbol{#1}}

\newcommand{\set}[1]{\mathbb{#1}}

\title{XQ-MEval: A Dataset with Cross-lingual Parallel Quality for Benchmarking Translation Metrics}

\author{
  \textbf{Jingxuan Liu$^\dagger$\textsuperscript{1}}\quad
  \textbf{Zhi Qu$^\dagger$\textsuperscript{1}}\quad
\\
  \textbf{Jin Tei\textsuperscript{1}}\quad
  \textbf{Hidetaka Kamigaito\textsuperscript{1}}\quad
  \textbf{Lemao Liu\textsuperscript{2}}\quad
  \textbf{Taro Watanabe\textsuperscript{1}}
\\
  $^\dagger$ These authors contributed equally to this work.
\\
  \textsuperscript{1}Nara Institute of Science and Technology, Japan.
\\
  \textsuperscript{2}Fudan University, China.
\\
  \href{mailto:jingxuan.liu.jm2@naist.ac.jp}{\texttt{jingxuan.liu.jm2@naist.ac.jp}}
}

\begin{document}
\maketitle
\begin{abstract}
Automatic evaluation metrics are essential for building multilingual translation systems.
The common practice of evaluating these systems is averaging metric scores across languages, yet this is suspicious since metrics may suffer from cross-lingual scoring bias, where translations of equal quality receive different scores across languages.
This problem has not been systematically studied because no benchmark exists that provides parallel-quality instances across languages, and expert annotation is not realistic.
In this work, we propose XQ-MEval, a semi-automatically built dataset covering nine translation directions, to benchmark translation metrics.
Specifically, we inject MQM-defined errors into gold translations automatically, filter them by native speakers for reliability, and merge errors to generate pseudo translations with controllable quality.
These pseudo translations are then paired with corresponding sources and references to form triplets used in assessing the qualities of translation metrics.
Using XQ-MEval, our experiments on nine representative metrics reveal the inconsistency between averaging and human judgment and provide the first empirical evidence of cross-lingual scoring bias.
Finally, we propose a normalization strategy derived from XQ-MEval that aligns score distributions across languages, improving the fairness and reliability of multilingual metric evaluation.\footnote{The code and dataset are available at: \url{https://github.com/zhiqu22/XQ-MEval}.}

\end{abstract}

\section{Introduction}
\label{sec:intro}
\begin{figure}[!ht]
    \centering
        \includegraphics[width=\linewidth]{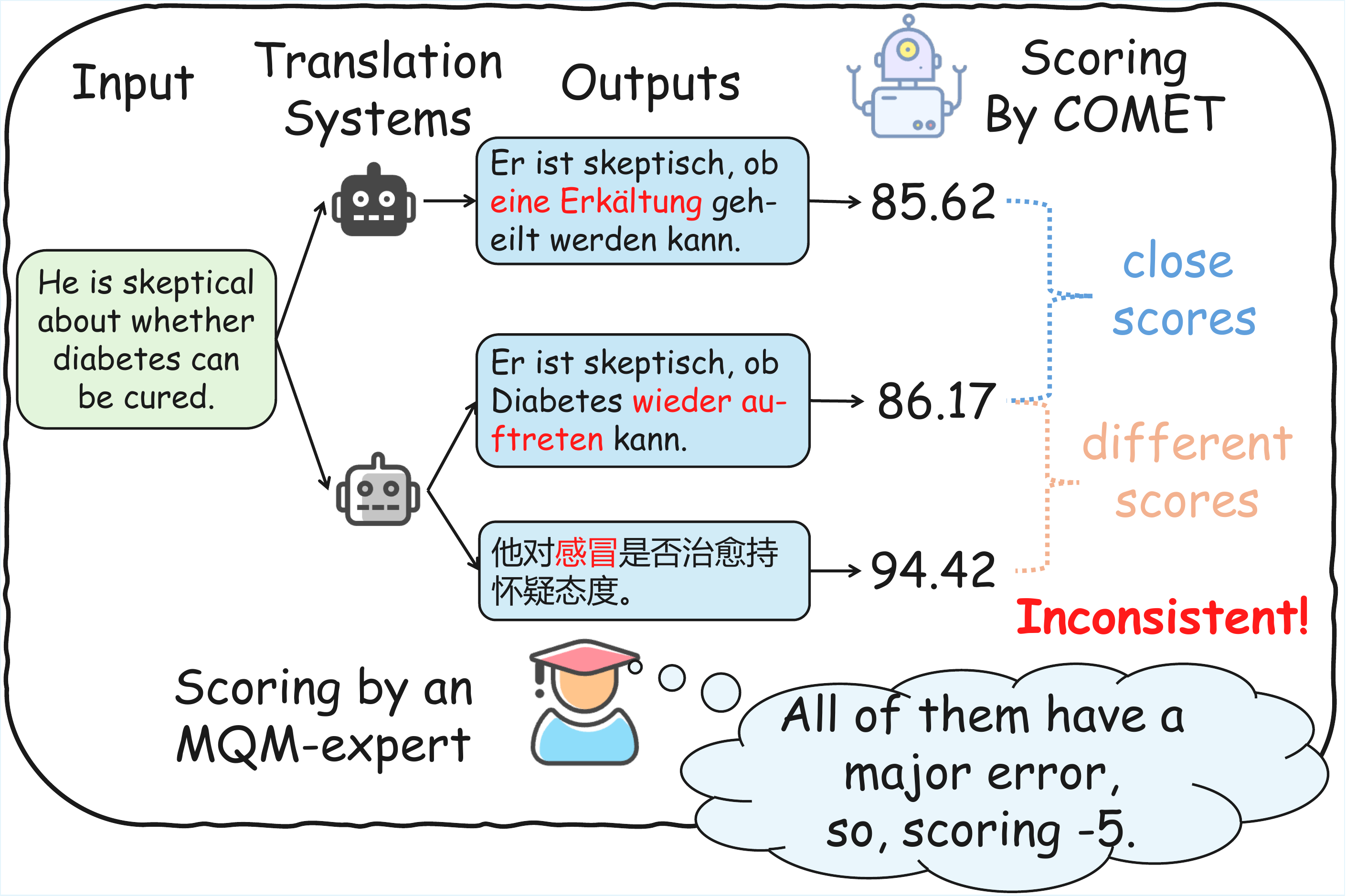}
    \caption{
    A clue of this study, showing the inconsistency between human evaluation, i.e., MQM, and automatic metrics, e.g., COMET.
    Three translations each contain one major error, thus sharing the same MQM score, yet COMET assigns notably different scores, with larger gaps across languages.
    }
    \label{fig:clue}
\end{figure}
With the growing demand for multilingual translation systems, comprehensive and reliable evaluation has become critical \cite{kocmi-etal-2024-findings}.
In human evaluation, Multidimensional Quality Metrics (MQM) largely achieves cross-lingually comparable evaluation through standardized error categories and hierarchical deduction~\cite{burchardt-2013-mqm,freitag-etal-2021-experts}.
However, as evaluation scales up, automatic evaluation metrics are essential due to their efficiency and scalability~\cite{popovic-2015-chrf,popovic-2017-chrf,post-2018-call,goyal-etal-2022-flores}.
Therefore, MQM driven automatic metrics have recently become the primary tools, e.g., COMET \cite{rei-etal-2020-comet} and MetricX~\cite{juraska-etal-2023-metricx}.

In multilingual translation evaluation, the common practice is to evaluate each language direction with a metric and then average the metric scores to compute a system-level score\footnote{The computational procedure of the average strategy is described in Appendix~\ref{appendix:pseudocode} with pseudocode.} \cite{chen-etal-2023-target,cao-etal-2024-exploring, qu-etal-2025-languages, qu-etal-2025-registering, qu-etal-2025-improving}.
However, this average strategy may be problematic because it implicitly assumes that different languages are scored on the same scale for a similar error.
In fact, cross-lingual scoring bias is indeed observed as illustrated in Figure \ref{fig:clue}.
To quantify and verify this potential problem, a benchmark is needed that provides parallel quality across languages, ensuring that cross-lingual comparisons are made on the same grounds, i.e., similar errors are quantified equally across different languages.
Due to the unaffordable cost of expert-level annotations, no such benchmark currently exists.

In this work, we propose a novel semi-automatic pipeline that injects MQM-defined errors into gold translations and filters them with native speakers, ensuring reliability and cross-lingual consistency.
By merging individual errors, we generate pseudo translations with controllable quality, which are then paired with gold sources and references to form triplets.
Based on this, we construct a dataset for evaluating metrics with cross-lingual parallel quality, namely XQ-MEval.
This dataset covers nine languages\footnote{Appendix \ref{appendix:language-selection} shows the details in language selection.}, i.e., Chinese, Japanese, Lao, Vietnamese, Indonesian, French, Spanish, Sinhala, and German, for translation directions from English, and provides parallel-quality triplets  for the fair metric comparisons across languages.

Based on XQ-MEval, we conduct experiments on nine representative automatic metrics.
The results reveal a clear inconsistency between averaging and human evaluation, and provide the first empirical evidence of cross-lingual scoring bias.
This bias has two manifestations: (1) systems of equal quality receive different scores across languages; (2) the decline of metric scores with decreasing quality is inconsistent across languages.
Building on this finding, we propose a simple strategy based on normalization \cite{book}, i.e., Language-specific Global Normalization (LGN), to calibrate multilingual evaluation metrics.
Our experiments show that, compared to the average strategy, LGN effectively reduces score range disparities and improves the fairness and reliability of multilingual metric evaluation.
We make the following threefold contributions in this study:
\begin{itemize}
    \item We present XQ-MEval, the first multilingual dataset with parallel-quality triplets across nine translation directions, enabling benchmarking of automatic evaluation metrics.
    \vspace{-0.5em}
    \item We evaluate representative metrics to reveal the inconsistency between the average strategy and human judgment, and provide the first analysis of cross-lingual scoring bias.
    \vspace{-0.5em}
    \item We introduce and verify LGN, a normalized average strategy that calibrates metrics in evaluating multilingual translation systems.
    \vspace{-0.5em}
\end{itemize}

\begin{figure*}[!ht]
    \centering
        \includegraphics[width=\linewidth]{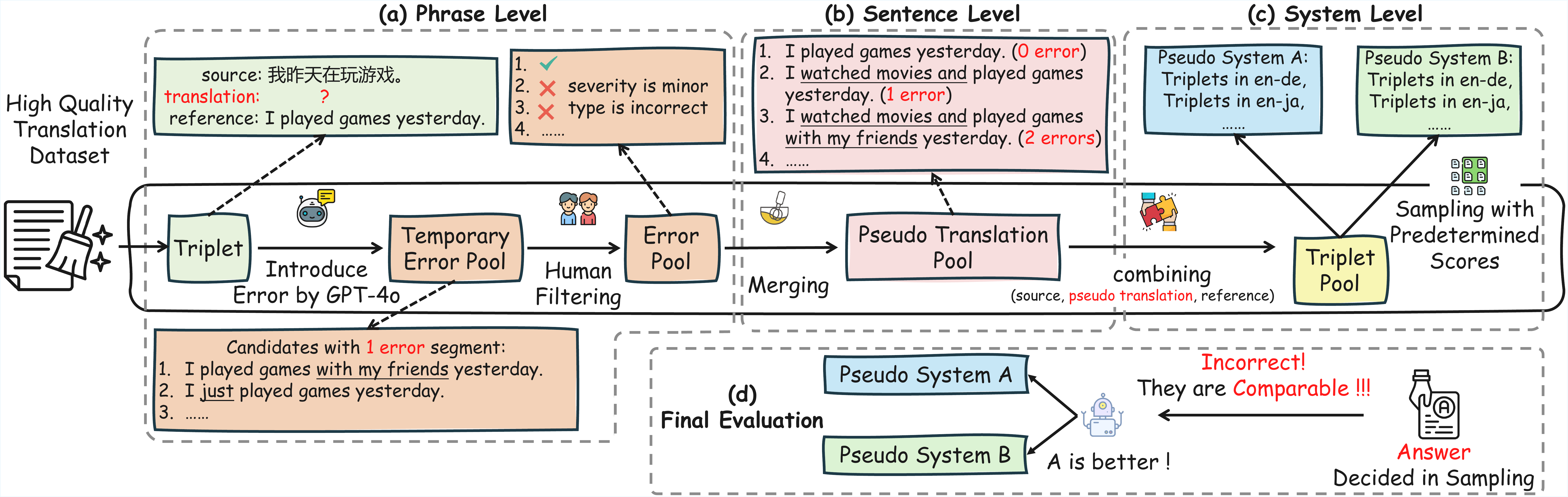}
    \caption{
    The illustration of our pipeline.
    Specifically, stages from (a) to (c) show the data construction and reveal that the product is to create pseudo translation systems with predetermined scores.
    Finally, stage (d) demonstrates the use of pseudo systems to assess the automatic metrics based on the answer, i.e., the predetermined score.
    }
    \vspace{-1em}
    \label{fig:pipeline}
\end{figure*}

\section{Related Work}\label{section:related_work}
The evaluation of bilingual translation systems relies on discrete scoring schemes \cite{koehn-2006-evaluate,vilar-2007-human,callison-burch-etal-2007-meta,denkowski-lavie-2010-choosing}, but these suffer from low inter-annotator agreement.
Although \citet{graham-etal-2013-continuous,bojar-2016-findings,bojar-2017-findings} introduced the continuous rating scale to mitigate this variability, subjectivity-related biases persisted across annotators.
Building upon the Multidimensional Quality Metrics (MQM) proposed by \citet{burchardt-2013-mqm}, \citet{freitag-etal-2021-experts} developed a framework that reduces annotator inconsistency through standardized error categories and hierarchical deduction.
Specifically, each sentence is assumed to have perfect quality initially, and points are deducted according to error type, e.g., accuracy and fluency, and severity, e.g., 1 for minor and 5 for major.
This makes translation metrics cross-lingually comparable because sentences with the same errors are expected to receive the same score across languages. 

To complement costly and inconsistent human-based evaluation, automatic evaluation metrics are proposed to approximate human judgments of translation quality efficiently.
They can be broadly categorized into three types: 
(1) \emph{Regression-based metrics} frame evaluation as a supervised task that directly predicts scalar quality scores, including both models trained explicitly for evaluation, e.g., COMET \cite{rei-etal-2020-comet,rei-etal-2022-comet,guerreiro-etal-2024-xcomet} and MetricX \cite{juraska-etal-2023-metricx, juraska-etal-2024-metricx}, and converting LLMs into evaluators, e.g., ReMedy \cite{tan-monz-2025-remedy}.
(2) \emph{Sequence-based metrics} evaluate translations by comparing candidate translations with gold references, primarily relying on surface-level similarity\footnote{Although metrics like BLEURT \cite{sellam-etal-2020-bleurt} are regression-based, the metric depended on embeddings from sequence information should be classified as sequence-based.}, e.g., BLEU \cite{papineni-etal-2002-bleu, post-2018-call} and chrF \cite{popovic-2015-chrf,popovic-2017-chrf}.
(3) \emph{Reference-free metrics}, also known as quality estimation (QE), extend regression-based methods to evaluate translations directly against the source without requiring references, e.g., COMET-kiwi \cite{rei-etal-2021-references,rei-etal-2023-scaling}.
In parallel, recent work has explored using LLMs as human evaluators by prompting them to follow explicit assessment agreements such as MQM, thereby approximating human judgment behavior at inference time \citep{kocmi-federmann-2023-gemba}.

These metrics are widely applied in multilingual translation evaluation, but the practice of averaging scores across languages \cite{share-2021,qu-watanabe-2022-adapting,chen-etal-2023-target,cao-etal-2024-exploring, qu-etal-2025-languages, qu-etal-2025-registering, qu-etal-2025-improving} may hinder the system-level evaluation since it is unclear whether a similar error is consistently measured across languages.
\citet{lyu2025minimumbayesriskdecoding} showed that, in error span detection, alignment with human judgments can vary with different decoding strategy.
Relatedly, \citet{von-daniken-etal-2025-measure} showed that metrics fail to align with human evaluation even in a single translation direction.
Thus, benchmarks are needed to expose cross-lingual scoring bias and guide metric improvement.
However, constructing them incurs costs similar to MQM, where each instance requires expert-level annotation.
Fortunately, using LLMs with human filtering can simplify this process \cite{li-etal-2023-halueval,kwan-etal-2024-mt,bai-etal-2024-mt,wang-etal-2025-ecomscriptbench}, providing a practical avenue for benchmark construction.

\section{Pipeline of Dataset Construction}
\label{subsec:data generation}
We present a multilingual dataset, XQ-MEval, for benchmarking automatic evaluation metrics covering nine translation directions, i.e., \texttt{en}-\texttt{zh}, \texttt{en}-\texttt{ja}, \texttt{en}-\texttt{lo}, \texttt{en}-\texttt{vi}, \texttt{en}-\texttt{id}, \texttt{en}-\texttt{fr}, \texttt{en}-\texttt{es}, \text{en}-\texttt{si}, and \texttt{en}-\texttt{de}, comprising both high-resource and low-resource languages\footnote{Languages are represented by ISO 639-1 codes, and details about language selection are shown in the Appendix~\ref{appendix:language-selection}.}.
Constructing such a dataset following MQM is challenging due to the high cost of expert annotation, which greatly limits the language coverage.
To address this, we employ a semi-automatic approach, formatting each sample as a triplet and rigorously controlling quality to ensure cross-lingual parallelism.
This design enables flexible sampling to simulate systems with predetermined quality levels for metric benchmarking.

Specifically, we introduce a novel pipeline for benchmark construction that enables systematic and cost-effective analysis of metric biases in Figure~\ref{fig:pipeline}, comprising phrase-level, sentence-level, and system-level stages of different granularity.
Automatic evaluation metrics operate on a triplet comprising a source, translation, and reference.
We begin with a high-quality translation corpus, where each translation pair forms the source and reference for a triplet.
At the phrase-level stage, a major-severity error is introduced into each reference.
Then, at the sentence-level stage, we merge 0 to 5 errors from such candidates\footnote{The choice of 5 follows Google’s MQM guideline, where each sentence can lose at most 25 points and each major error accounts for 5 points \cite{freitag-etal-2021-experts}.} to generate pseudo translations\footnote{Annotators' feedback indicates that although combining errors may appear unnatural, they remain objectively valid.} with six distinct quality levels.
Finally, at the system-level stage, pseudo systems are constructed by assembling triplets across different quality levels, thereby emulating translation systems with predetermined performance.

Nevertheless, we acknowledge that XQ-MEval instances are synthesized rather than produced by real translation systems, and may thus differ from real-world scenarios.
We have conducted preliminary experiments on usable real-world MQM datasets and validated our approach in Appendix~\ref{appendix:preliminary experiment}.

\begin{table*}[!ht]
    \centering
    \resizebox{0.99\linewidth}{!}{
    \begin{tabular}{cp{0.15\linewidth}p{0.3\linewidth}cp{0.6\linewidth}p{0.25\linewidth}} 
    \toprule
        \textbf{Part} & \multicolumn{1}{c}{\textbf{Product}} & \multicolumn{1}{c}{\textbf{Operation}} & \textbf{Language} & \multicolumn{1}{c}{\textbf{Example}} & \multicolumn{1}{c}{\textbf{Note}}\\
    \midrule
        \multirow{13}{*}{1} & \multirow{13}{=}{\centering Error Pool} & \multirow{13}{=}{\centering Introducing single error to the reference by GPT-4o, then filter by native speakers.} 
        & \multirow{4}{*}{\centering \texttt{de}} & Der Klage zufolge wurde der Abfall aus dem UN-Lager nicht ordnungsgemäß gesäubert, was dazu führte, dass Bakterien in den Zufluss des Artibonit-Flusses, einem der größten Flüsse Haitis, \textcolor{red}{<v>sowie in andere Gewässer</v>} gelangten.  & \multirow{4}{=}{\centering Error type: Addition;\\ Human judgment: \greencheck} \\
        \cdashline{4-6}
        &&& \multirow{3}{*}{\centering \texttt{ja}} & \jp{訴訟によれば、国連キャンプの廃棄物が適切に消毒されていなかったため、ハイチ最大級の\textcolor{red}{<v></v>}に細菌が侵入したとのことです。}& \multirow{3}{=}{\centering Error type: Omission;\\ Human judgment: \greencheck} \\
        \cdashline{4-6}
        &&& \multirow{3}{*}{\centering \texttt{zh}} & \zh{诉讼材料显示，联合国营地未能对废弃物进行\textcolor{red}{<v>彻底的焚烧</v>}，因而导致细菌进入阿蒂博尼特河的支流，这是海地最大河流之一。} & \multirow{3}{=}{\centering Error type: Mistranslation;\\ Human judgment: \greencheck}\\ 
        \cdashline{4-6}
        &&& \multirow{3}{*}{\centering \texttt{zh}} & \zh{诉讼材料显示，联合国营地未能对废弃物进行适当的消毒处理，因而导致细菌进入\textcolor{red}{<v>Artibonite River</v>}的支流，这是海地最大河流之一。}& \multirow{3}{=}{\centering Error type: Untranslated;\\ Human judgment: \redcross} \\
    \midrule  
        \multirow{10}{*}{2} & \multirow{10}{=}{\centering Pseudo Translation Pool} & \multirow{10}{=}{\centering Merging several candidates, where the error span is not conflict, to get a pseudo translation.}
        & \multirow{3}{*}{\centering \texttt{ja}} & \jp{訴訟によれば、国連キャンプの\textcolor{red}{<v>食料供給</v>}が適切に消毒されていなかったため、ハイチ最大級のアルティボナイト川の支流に細菌が侵入したとのことです。}& \multirow{3}{=}{\centering Error span: 14-18} \\
        \cdashline{4-6}
        &&& \multirow{4}{*}{\centering \texttt{ja}} & \jp{訴訟によれば、国連キャンプの廃棄物が適切に消毒されていなかったため、\textcolor{red}{<v>さらに多くの問題が発生し、</v>}ハイチ最大級のアルティボナイト川の支流に細菌が侵入したとのことです。}& \multirow{4}{=}{\centering Error span: 34-35} \\
        \cdashline{4-6}
        &&& \multirow{4}{*}{\centering \texttt{ja}} & \jp{訴訟によれば、国連キャンプの\textcolor{red}{<v>食料供給</v>}が適切に消毒されていなかったため、 \textcolor{red}{<v>さらに多くの問題が発生し、</v>}ハイチ最大級的阿蒂博尼特川的支流に細菌が侵入したとのことです。}& \multirow{4}{=}{\centering Pseudo Translation\\ with 2 Errors} \\
    \midrule
        \multirow{10}{*}{3} & \multirow{10}{=}{\centering Triplet Pool} & \multirow{10}{=}{\centering Each triplet fed into metrics is combined by a source, a translation, and a reference.}
        & \multirow{3}{*}{\centering \texttt{en}} & According to the lawsuit, waste from the UN camp was not properly sanitized, causing bacteria to enter the tributary of the Artibonite River, one of Haiti's largest. & \multirow{3}{=}{\centering Source} \\
        \cdashline{4-6}
        &&& \multirow{4}{*}{\centering \texttt{ja}} & \jp{訴訟によれば、国連キャンプの\textcolor{red}{<v>食料供給</v>}が適切に消毒されていなかったため、 \textcolor{red}{<v>さらに多くの問題が発生し、</v>}ハイチ最大級的阿蒂博尼特川的支流に細菌が侵入したとのことです。}& \multirow{4}{=}{\centering Pseudo Translation} \\
        \cdashline{4-6}
        &&& \multirow{3}{*}{\centering \texttt{ja}} & \jp{訴訟によれば、国連キャンプの廃棄物が適切に消毒されていなかったため、ハイチ最大級的阿蒂博尼特川的支流に細菌が侵入したとのことです。} & \multirow{3}{=}{\centering Reference} \\
    \bottomrule
    \end{tabular}
    }
    \caption{Examples used to assist in explaining Figure \ref{fig:pipeline}.
    The column of part is used for conveniently referring.}
    \label{tab:detailed_pipeline}
\end{table*}

\subsection{Phrase-level Construction}
\label{subsuc:phrase construction}
XQ-MEval is built on Flores\footnote{\url{https://huggingface.co/datasets/openlanguagedata/flores_plus}}, a high-quality multilingual translation dataset, denoted as \(\set{F}\), with 102 instances used in our experiments\footnote{We have manually selected to exclude very short sentences that cannot accommodate multiple injected errors.}.
Flores is particularly suitable because its translations are semantically parallel and are carefully validated by multiple native speakers \cite{nllb}.

As shown in Figure \ref{fig:pipeline}, we define each translation instance in \(\set{F}\) as $(s, r)$ where $s$ represents the source in \texttt{en} and $r$ represents its reference.
We employ GPT-4o\footnote{Version: \textit{gpt-4o-2024-11-20}.} \cite{gpt4} to inject an MQM-defined error of major severity into $r$, producing a temporary error candidate $\hat{r}$ comprising a single error segment with an identification tag.

We introduce the following four error types, which dominate existing MQM datasets\footnote{These four types account for 46.3\% of all MQM errors.} and are conducive to cross-lingual comparability as they are purely semantic \cite{45323cee-e552-322a-8a59-c8ec5d452b38,cristofaro2009grammatical}:
(1) \emph{Addition}, where extraneous information is inserted in translations;
(2) \emph{Omission}, where a part of the source is left out;
(3) \emph{Mistranslation}, where the meaning is distorted or incorrect;
(4) \emph{Untranslated}, where the source remains untranslated text.
Because each pseudo translation $\Tilde{r}$ may contain up to five errors in our settings, we allow multiple instances of the same error type injected separately into the first and second halves of the sentence, which are first divided and explicitly tagged to guide GPT-4o to introduce error segments into the corresponding parts.
Thus, a single $(s, r)$ can yield up to eight temporary error candidates $\ve{\hat{r}} = \{\hat{r}_1,\hat{r}_2,\ldots,\hat{r}_8\}$.
Applying this process to the entire dataset produces a temporary error pool \(\set{\hat{R}} = \bigcup_{i=1}^n \ve{\hat{r}}_i\).\footnote{Prompts are carefully designed and listed in Appendix \ref{appendix:prompt}.}

Then, native speakers of the nine target languages review and filter \(\set{\hat{R}}\).
In practice, two independent reviewers are engaged, but, for \texttt{si}, \texttt{lo}, and \texttt{vi}, only one reviewer is available due to resource constraints.
Finally, only $\hat{r}$ unanimously approved by both annotators are retained to construct the final error pool \(\set{\hat{R}_\text{filtered}}\).
The part 1 of Table \ref{tab:detailed_pipeline} demonstrates this process.

\begin{table}[t]
    \centering
     \resizebox{\linewidth}{!}{
    \begin{tabular}{cccccccccc}
    \toprule
        & \texttt{en}-\texttt{zh} & \texttt{en}-\texttt{lo} & \texttt{en}-\texttt{ja} & \texttt{en}-\texttt{vi} & \texttt{en}-\texttt{id} & \texttt{en}-\texttt{fr} & \texttt{en}-\texttt{es} & \texttt{en}-\texttt{si} & \texttt{en}-\texttt{de} \\ 
    \midrule
        Add. & 194 & 196 & 194 & 201 & 196 & 200 & 196 & 200 & 203 \\ 
        Omit. & 200 & 191 & 196 & 199 & 197 & 196 & 197 & 188 & 194 \\ 
        Mist. & 201 & 200 & 202 & 200 & 201 & 196 & 197 & 197 & 194 \\ 
        Untr. & 181 & 172 & 183 & 171 & 188 & 183 & 181 & 181 & 183 \\ 
    \bottomrule
    \end{tabular}
    }
    \caption{The number of candidates generated by GPT-4o and filtered by annotators for each error type. The abbreviations of error type are as follows: Addition, Omission, Mistranslation, and Untranslated.}
    \label{tab:filter_statistics}
\end{table}

To ensure consistency, we provide detailed annotation guidelines in Appendix~\ref{appendix:instruction} that explain the four MQM errors and specify filtering conditions regarding completeness, locality, and severity.
Table \ref{tab:filter_statistics} summarizes the number of sentences generated by GPT-4o and retained by annotators for each error type.
Also, to assess annotation reliability, we compute inter-annotator agreement between the two native speakers.
As shown in Table \ref{tab:human_alignment}, agreement is consistently high, reflecting the effectiveness of our guidelines.
We further validate robustness through a second round of independent screening on 200 randomly sampled \texttt{en}-\texttt{zh} and \texttt{en}-\texttt{ja} instances.
The alignment rates between the two rounds are 99\% for \texttt{en}-\texttt{zh} and 98\% for \texttt{en}-\texttt{ja}, confirming the stability of annotation process.
These results demonstrate that the constructed dataset is both reliable and reproducible, establishing a solid foundation for subsequent stages.
\begin{table}[t]
    \centering
    \resizebox{0.9\linewidth}{!}{
    \begin{tabular}{lllllll}
    \toprule
         & \texttt{en}-\texttt{zh} & \texttt{en}-\texttt{ja} & \texttt{en}-\texttt{fr} & \texttt{en}-\texttt{es} & \texttt{en}-\texttt{de} & \texttt{en}-\texttt{id} \\
    \midrule
        Agreement (\%) & 98.16 & 96.45 & 97.79 & 97.30 & 97.67 & 96.45 \\
    \bottomrule
    \end{tabular}
    }
    \caption{The annotation agreement between the two native speakers during the manual screening process.}
    \label{tab:human_alignment}
\end{table}
\subsection{Sentence-level Construction}
\begin{table}[t]
    \centering
     \resizebox{\linewidth}{!}{   
    \begin{tabular}{crrrrrrrrr}
    \toprule
        & \texttt{en}-\texttt{zh} & \texttt{en}-\texttt{lo} & \texttt{en}-\texttt{ja} & \texttt{en}-\texttt{vi} & \texttt{en}-\texttt{id} & \texttt{en}-\texttt{fr} & \texttt{en}-\texttt{es} & \texttt{en}-\texttt{si} & \texttt{en}-\texttt{de}\\
    \midrule
        max & 176 & 176 & 150 & 218 & 176 & 139 & 139 & 139 & 176 \\
        min & 19 & 8 & 11 & 21 & 7 & 8 & 11 & 11 & 15 \\ 
    \bottomrule
    \end{tabular}
    }
    \caption{Summarizes the maximum and minimum number of pseudo translations generated for each triplet in different translation directions.}
    \label{tab:max_and_min}
\end{table}

Based on \(\set{\hat{R}_\text{filtered}}\), we generate each pseudo translation $\Tilde{r}$ by merging $k$ single-error candidates $\hat{r}$, where $k \in \{0, 1, 2, 3, 4, 5\}$, all of which are from the same $\ve{\hat{r}}_\text{filtered}$, i.e., the candidates filtered for each pair $(s,r)$, as illustrated in Figure \ref{fig:pipeline}.
$\Tilde{r}$ is a variant of $r$ containing between 0 and 5 errors, thus covering six distinct quality levels in the MQM framework.
Part 2 of Table \ref{tab:detailed_pipeline} provides an example, where two non-overlapping $\hat{r}$ are merged to form a $\Tilde{r}$ with two errors.
In addition, a special case is that of 0 error, corresponding to the reference itself.\footnote{In this case, a metric should assign a full score to the triplet, when the translation matches the gold reference.}

By merging candidates, we can flexibly produce pseudo translations with the desired scores.
However, candidates may contain overlapping error spans, which compromise the locality of each error.
Such overlapping combinations are simply discarded so that the actual number of pseudo translations is smaller than the theoretical maximum.
As a result, each triplet yields a set of pseudo translations that cover different quality levels.
Table~\ref{tab:max_and_min} reports the minimum and maximum number of pseudo translations generated per triplet for each language direction, reflecting the constraints imposed by overlap and sentence structure.

\subsection{System-level Construction and Final Evaluation}\label{subsec:data generation system-level}
As shown in part 3 of Table \ref{tab:detailed_pipeline}, an instance is formed as a triplet $(s,\Tilde{r},r)$.
By iterating over the entire dataset, we obtain the triplet pool $\mathcal{D}$, which constitutes the final dataset of XQ-MEval.

Figure \ref{fig:pipeline} further illustrates how $\mathcal{D}$ enables systematic benchmarking of automatic metrics.
We assume the existence of a translation system with a given MQM score derived from the number of error spans and then construct a pseudo system by sampling triplets that reflect this target performance.
This procedure is both flexible and powerful because it allows us to generate arbitrary pseudo systems tailored to different evaluation scenarios. 
Based on pseudo systems with predefined performance, we evaluate them using automatic metrics and measure the alignment between metric scores and predefined scores as a proxy for consistency with human judgments.\footnote{Appendix~\ref{appendix:pseudocode} exhibits the process of computing system-level metric scores, and shows comparing them to predefined scores, i.e., human evaluations.}

\section{Experimental Setup}
Based on XQ-MEval in Section~\ref{subsec:data generation}, we perform a large-scale and multilingual analysis of existing automatic evaluation metrics\footnote{We primarily focus on metrics within the categories defined in Section~\ref{section:related_work}. However, we also analyze LLM-based approaches, including LLM-adapted regression metrics and MQM-style LLM-as-judge evaluation, in Appendix~\ref{appendix:llm-as-judge}.} as follows.

\begin{table*}[!ht]
\centering

\begin{subtable}{0.48\linewidth}
\centering
\resizebox{\linewidth}{!}{
\begin{tabular}{cccccccc}
\toprule
\multirow{2}{*}[-0.6ex]{\shortstack{Num. of \\ Lang.}} & \multicolumn{7}{c}{System-level Kendall-$\tau$} \\ 
\cmidrule(lr){2-8}
 & BLR. & COM. & xCOM. & MX-r. & KW22. & KW23. & MX-q.\\ 
\midrule
3 & 0.89 & 0.88 & 0.90 & 0.80 & 0.88 & 0.86 & 0.82 \\ 
6 & 0.88 & 0.88 & 0.89 & 0.84 & 0.90 & 0.89 & 0.83 \\ 
9 & 0.87 & 0.89 & 0.90 & 0.83 & 0.90 & 0.89 & 0.83  \\ 
\bottomrule
\end{tabular}
}
\caption{System-level}
\label{tab:a}
\end{subtable}
\hfill
\begin{subtable}{0.48\linewidth}
\centering
\resizebox{\linewidth}{!}{
\begin{tabular}{cccccccc}
\toprule
\multirow{2}{*}[-0.6ex]{\shortstack{Num. of \\ Lang.}} & \multicolumn{7}{c}{Triplet-level Kendall-$\tau$} \\ 
\cmidrule(lr){2-8}
 & BLR. & COM. & xCOM. & MX-r. & KW22. & KW23. & MX-q. \\ 
\midrule
 3 & 0.50 & 0.46 & 0.38 & 0.35 & 0.44 & 0.39 & 0.32 \\ 
 6 & 0.46 & 0.44 & 0.42 & 0.38 & 0.45 & 0.39 & 0.33 \\ 
 9 & 0.48 & 0.42 & 0.44 & 0.38 & 0.44 & 0.38 & 0.32 \\ 
\bottomrule
\end{tabular}
}
\caption{Triplet-level}
\label{tab:b}
\end{subtable}

\caption{Results showing the system-level and triplet-level Kendall-\(\tau\) correlation between averaged metric scores and human judgments on pseudo systems.
    Num. of Lang. denotes the number of involved languages.
    In this setting, Num. of 3 means that the system is sampled from \texttt{zh}, \texttt{lo}, and \texttt{de}; Num. of 6 means that the system is sampled from \texttt{zh}, \texttt{lo}, \texttt{de}, \texttt{id}, \texttt{ja}, and \texttt{si}; Num. of 9 means that the system is sampled from all languages.
    The abbreviations of metric are as follows: BLEURT, COMET, xCOMET, MX-reg, KIWI22, KIWI23, and MX-qe.}
\label{tab:original correlation}
\end{table*}

\paragraph{Sequence-based}
(1) spBLEU \cite{goyal-etal-2022-flores}, a variant of BLEU that unifies tokenization across languages through a SentencePiece tokenizer \cite{kudo-richardson-2018-sentencepiece};
(2) chrF++ \cite{popovic-2017-chrf}, which assesses character-level overlap and balances precision with recall;
(3) BLEURT-20 \cite{sellam-etal-2020-bleurt}, a BERT-based metric trained on human-annotated data to better align with human judgments.

\paragraph{Regression-based}
(1) COMET-22 \cite{rei-etal-2022-comet}, which integrates source, hypothesis, and reference embeddings to predict quality scores;
(2) xCOMET-XL \cite{guerreiro-etal-2024-xcomet}, which improves interpretability by detecting errors explicitly;
(3) MetricX-23 \cite{juraska-etal-2023-metricx}, abbreviated as MX-reg, initialized with mT5 \cite{xue-etal-2021-mt5} and fine-tuned on MQM data.

\begin{table*}[!t]
    \centering
    \resizebox{0.75\linewidth}{!}{
    \begin{tabular}{crrrrrrrrr}
    \toprule
         Quality Level& spBLEU & chrF & BLEURT & COMET & xCOMET & MX-reg & KIWI22 & KIWI23 & MX-qe \\
         \cmidrule(lr){1-1}\cmidrule(lr){2-4} \cmidrule(lr){5-7} \cmidrule(lr){8-10}
        1 & 1.53 & 7.56 & 1.62 & 2.51 & 9.95 & 22.80 & 2.38 & 6.05 & 23.61\\ 
        2 & 3.16 & 9.84 & 4.46 & 3.84 & 14.77 & 23.74 & 2.39 & 8.55 & 22.38 \\ 
        3 & 5.04 & 12.00 & 7.26 & 5.09 & 20.51 & 23.75 & 2.66 & 11.23 & 20.57 \\ 
        4 & 7.25 & 14.22 & 10.07 & 6.27 & 26.01 & 24.23 & 3.29 & 14.42 & 18.94 \\ 
        5 & 10.01 & 16.23 & 13.79 & 7.61 & 28.67 & 24.03 & 3.76 & 18.86 & 15.40 \\
        \bottomrule
    \end{tabular}
    }
    \caption{Illustration of the cross-lingual CV (\%) of scores for nine
automatic metrics measured at five quality levels.}
    \label{tab:cross-linguistic variance}
\end{table*}

\begin{figure*}[!t]
    \centering
        \includegraphics[scale=0.27]{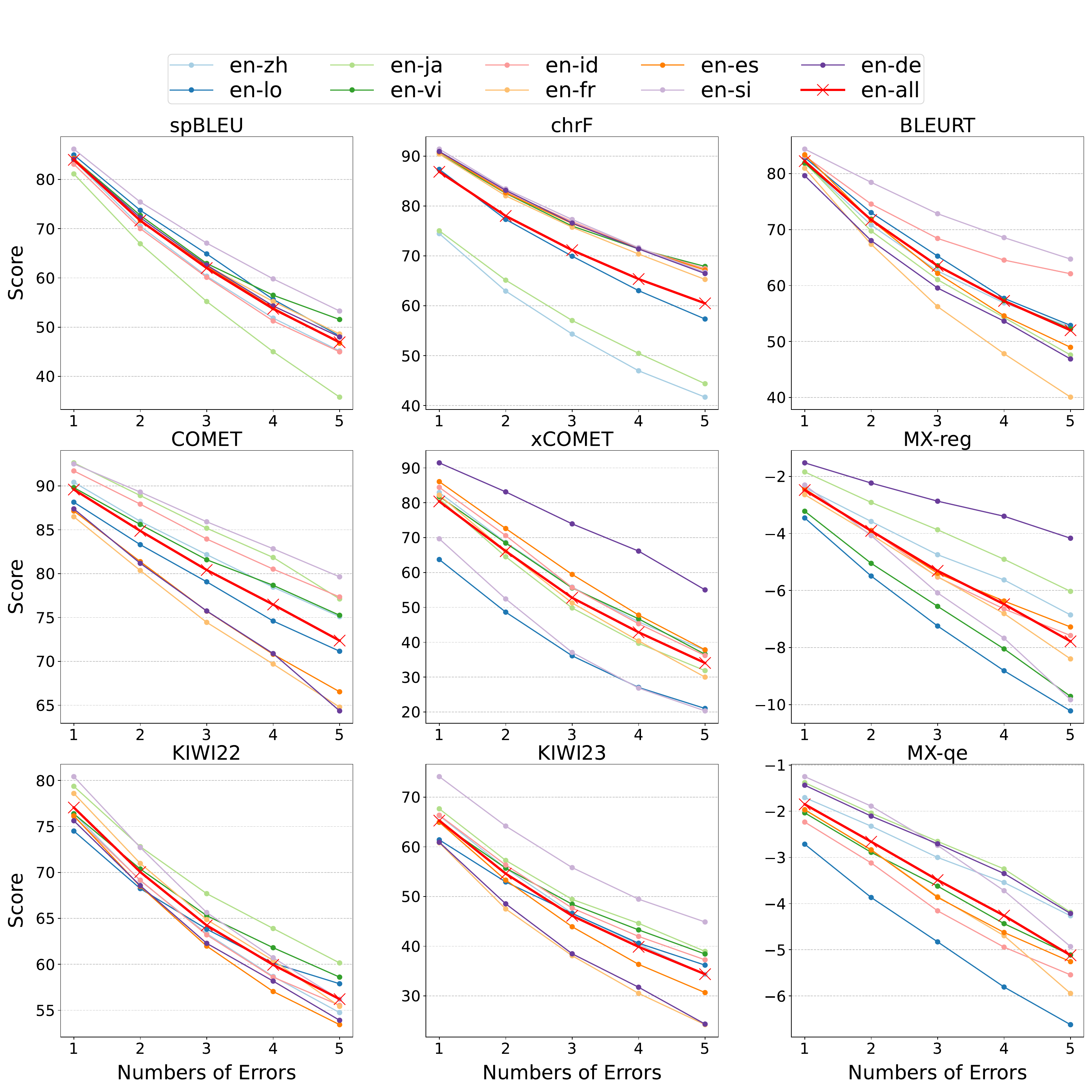}
    \caption{Visualization of nine metric scores across nine directions at varying translation quality levels.
    \texttt{en}-\texttt{all} denoting the average metric scores among all directions.}
    \label{fig:scores_graph}
\end{figure*}

\paragraph{Reference-free}
(1) COMET-KIWI-22 \cite{rei-etal-2022-cometkiwi}, abbreviated as KIWI22, a reference-free variant of COMET-22;
(2) COMET-KIWI-23 \cite{rei-etal-2023-scaling}, abbreviated as KIWI23, an extended version of KIWI22;
(3) MetricX-23-QE \cite{juraska-etal-2023-metricx}, abbreviated as MX-qe, the reference-free variant of MetricX-23.

\section{Analysis on Average Strategy}
\subsection{Verification}\label{subsec:verification}
To verify the consistency between the average strategy and human evaluations in multilingual MT evaluation, we assemble 10 pseudo systems to approximate real-world translation systems.

Following the procedure of Section~\ref{subsec:data generation system-level}, each pseudo system is built by aggregating 102 triplets sampled per language pair from multiple languages to meet predetermined scores.
After scoring each triplet, system-level metric scores are computed by averaging their respective scores across directions, followed by calculating their correlation with human evaluation to assess agreement.
This procedure is repeated 100 times for stability, and the average correlation across these repetitions is reported.
We rely on the Kendall-\(\tau\) coefficient \cite{kendall1938new}, a statistical measure of rank correlation, to quantify the consistency between the rankings induced by metrics and by predetermined scores, where higher values indicate stronger consistency and vice versa.

Table~\ref{tab:original correlation}(\subref{tab:a}) reports the system-level correlation results under three settings with 3, 6, and all 9 languages, where the subsets of 3 and 6 were selected to maximize linguistic diversity.
Although correlations appear high across settings, this is expected in our simplified evaluation setup, where instance quality is divided into five coarse-grained levels with large gaps, making quality differences easier for metrics to distinguish.
As a result, such high correlations may be inflated by the evaluation setup and should be interpreted with caution.

To further examine whether this apparent consistency holds at a finer granularity, we analyze metric behavior at the triplet level.
Since pseudo systems are constructed from triplets, we group all possible triplets across languages to form test systems.
Table~\ref{tab:original correlation}(\subref{tab:b}) presents the resulting triplet-level correlations, which are substantially lower and indicate pronounced inconsistency.
These results shed light on the concerns raised by the system-level analysis and point to potential cross-lingual inconsistencies in metric scoring behavior.

\subsection{Analysis}\label{subsec:analysis}
To analyze inconsistencies between metrics and human evaluations, we construct pseudo monolingual systems, each restricted to a single translation direction and quality level.
Unlike multilingual systems, this setting isolates metric behavior within one language and enables direct cross-language comparison at the same quality level.
Moreover, to address imbalances in triplet counts across quality levels\footnote{Appendix~\ref{appendix:numbers of sentences of 5 levels} counts and lists the triplets distribution of different languages.}, we randomly sample 102 triplets per system and repeat this procedure 10 times to ensure robustness.\footnote{We further report tests with 5, 10, and 25 repetitions in Appendix~\ref{appendix: mean and variance for 5, 10, 25 sampling} to support our design choices.}

\paragraph{At the same quality level}
Table~\ref{tab:cross-linguistic variance} reports cross-lingual coefficients of variation (CV) for nine metrics across five quality levels, corresponding to translations with the number of errors ranging from 1 to 5.
For each quality level, CV is computed from the mean and standard deviation of metric scores across nine monolingual systems.
CV measures score inconsistency across languages at the same quality level, indicating whether metrics provide consistent judgments as translation direction varies, with ideal values close to zero.
Results show inconsistencies for most metrics, with CV increasing as translation quality decreases.
This indicates that metrics assign divergent scores to translations of comparable quality, deviating from human evaluation and reflecting cross-lingual bias in the scoring behavior of metrics.

\begin{table*}[!ht]
\centering

\begin{subtable}{0.48\linewidth}
\centering
\resizebox{\linewidth}{!}{
\begin{tabular}{cccccccc}
\toprule
\multirow{2}{*}[-0.6ex]{\shortstack{Num. of \\ Lang.}} & \multicolumn{7}{c}{System-level Kendall-$\tau$} \\ 
\cmidrule(lr){2-8}
 & BLR. & COM. & xCOM. & MX-r. & KW22. & KW23. & MX-q.\\ 
\midrule
3 & \textbf{0.90} & \textbf{0.89} & \textbf{0.91} & \textbf{0.88} & \textbf{0.90} & \textbf{0.88} & \textbf{0.85} \\ 
6 & \textbf{0.92} & \textbf{0.91} & \textbf{0.90} & \textbf{0.91} & \textbf{0.92} & \textbf{0.90} & \textbf{0.86} \\ 
9 & \textbf{0.91} & \textbf{0.93} & \textbf{0.92} & \textbf{0.88} & \textbf{0.91} & \textbf{0.91} & \textbf{0.86} \\ 
\bottomrule
\end{tabular}
}
\caption{System-level}
\end{subtable}
\hfill
\begin{subtable}{0.48\linewidth}
\centering
\resizebox{\linewidth}{!}{
\begin{tabular}{cccccccc}
\toprule
\multirow{2}{*}[-0.6ex]{\shortstack{Num. of \\ Lang.}} & \multicolumn{7}{c}{Triplet-level Kendall-$\tau$} \\ 
\cmidrule(lr){2-8}
 & BLR. & COM. & xCOM. & MX-r. & KW22. & KW23. & MX-q. \\ 
\midrule
3 & \textbf{0.51} & \textbf{0.48} & \textbf{0.47} & \textbf{0.41} & \textbf{0.45} & \textbf{0.41} & \textbf{0.34} \\ 
6 & \textbf{0.49} & \textbf{0.48} & \textbf{0.48} & \textbf{0.42} & \textbf{0.46} & \textbf{0.41} & \textbf{0.35} \\ 
9 & \textbf{0.50} & \textbf{0.48} & \textbf{0.49} & \textbf{0.41} & \textbf{0.45} & \textbf{0.40} & \textbf{0.34} \\ 
\bottomrule
\end{tabular}
}
\caption{Triplet-level}
\end{subtable}

\caption{Kendall-\(\tau\) correlations at system-level and triplet-level, corresponding to Table~\ref{tab:original correlation}.
    All settings and abbreviations follow Table~\ref{tab:original correlation}.
    Bold values indicate improvements of LGN over the average strategy. 
    Improvements are modest in magnitude but statistically significant; significance tests are reported in Appendix~\ref{appendix:significance test}.}
\label{tab:normalized correlation}
\end{table*}

\paragraph{Across different quality level}
Figure~\ref{fig:scores_graph} plots metric scores across translation directions at varying quality levels to examine whether score trends remain consistent as quality varies.\footnote{Specific values are provided in Appendix~\ref{appendix:graphs and detailed scores}.}
Curves across directions should overlap, with similar scores and trends across quality levels.
In contrast, two phenomena are observed.\footnote{Appendix~\ref{appendix:detailed explanation for score decline} describes the difference across directions and across metrics in detail.}
First, metric scores differ across directions even at the same quality level.
Second, as quality decreases, score reduction rates vary across directions, leading to widening gaps between curves.
Consistent with the analysis in Table~\ref{tab:cross-linguistic variance}, these variations confirm the existence of cross-lingual scoring bias in automatic translation metrics, posing a challenge for metrics to align with human evaluations in multilingual settings, where uniformity across directions is expected.

\section{Normalization-based Scoring}
\subsection{Methodology}
\label{subsec:methodology}
The analysis in Section~\ref{subsec:analysis} reveals substantial variation in metric score ranges across translation directions.
Figure~\ref{fig:comet} further illustrates this issue using COMET, where the distribution of scores for different target languages diverges even when the human score is fixed at 15, comprising 3 errors in each translation.
It is evident that different languages occupy distinct numerical scales, making metric scores inconsistent even when human quality is comparable.

To address this problem, we propose \textbf{Language-specific Global Normalization} (LGN), which adopts z-score normalization to unify score scales across languages via mean and standard deviation.
LGN computes the mean and standard deviation of triplet scores for each translation direction across all quality levels.
For a given direction, 102 triplets are randomly sampled per quality level (including error-free translations) and pooled to calculate the global mean and standard deviation.\footnote{Appendix~\ref{appendix:pseudocode} describes the computational procedure with pseudo codes.}
This process is repeated 10 times, and the final values are obtained by averaging across repetitions.
By normalizing scores, LGN effectively reduces discrepancies between score ranges by narrowing the gaps in score distributions.
The general formula for normalization is as follows, with $\mu$ and $\sigma$ being the direction-wise mean and standard deviation:
\begin{figure}[!t] 
    \centering
        \includegraphics[width=\linewidth]{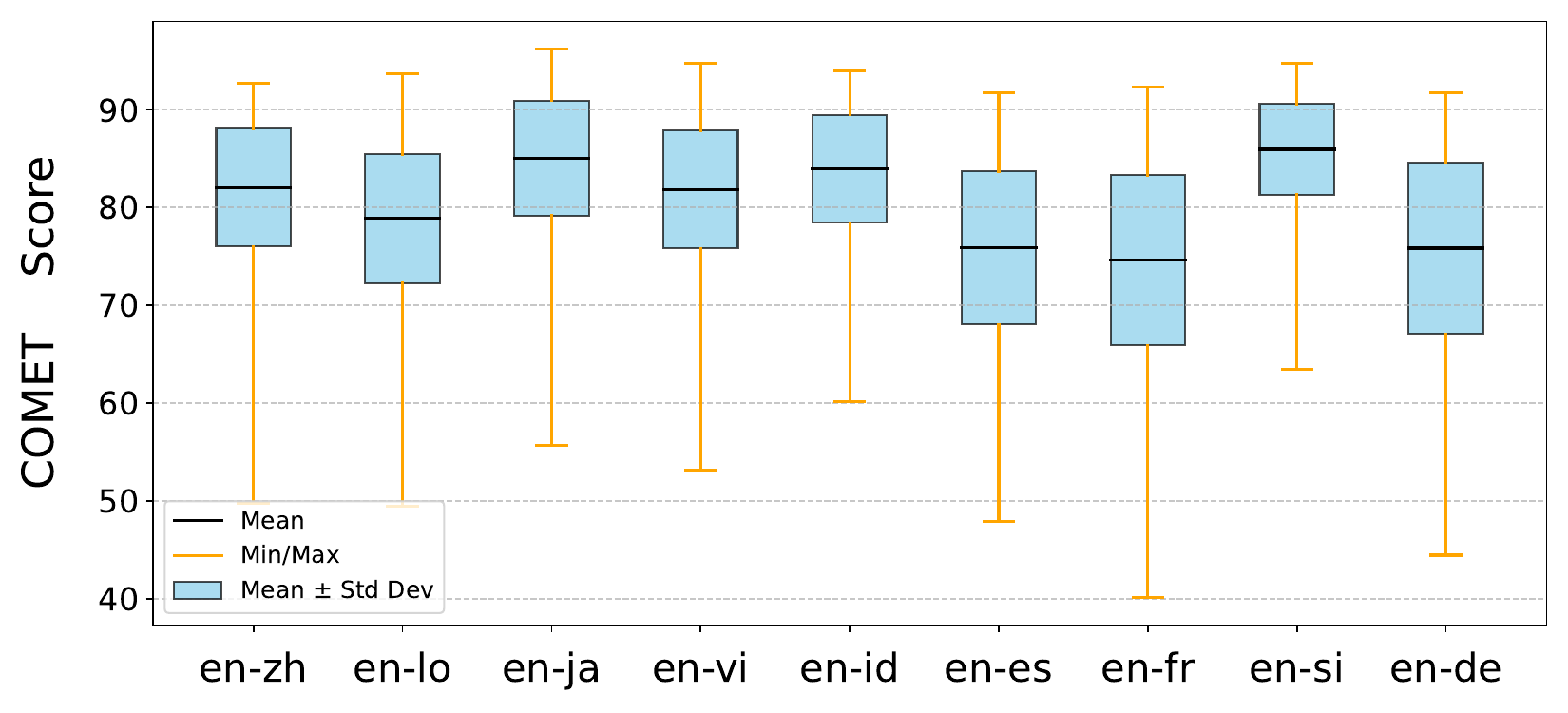}
    \caption{The illustration of COMET score distribution across different translation directions under fixed human evaluation scores.
    The bar sections represent the mean $\pm$ standard deviation, while the whiskers indicate the maximum and minimum values.
    }
    \label{fig:comet}
\end{figure}
\begin{equation}
z = \frac{\text{score} - \mu}{\sigma}.
\end{equation}

\subsection{Experiments and Results}
We evaluate LGN by applying it before cross-lingual score averaging, following the same experimental setup as in Table~\ref{tab:original correlation}.
Results in Table~\ref{tab:normalized correlation} show that LGN consistently improves the correlation between automatic metrics and human evaluations in multilingual settings.
Although the absolute gains are moderate, partly because correlations are already high under the original setup, paired-sample t-tests reported in Appendix~\ref{appendix:significance test} confirm the statistically-significant improvement.
Also, this reflects the concern raised in the system-level verification of Section \ref{subsec:verification}, where the value shown in Table \ref{tab:original correlation}(\subref{tab:a}) is high but still suboptimal due to the cross-lingual scoring bias.
By reducing disparities in score ranges, LGN improves cross-lingual consistency both the system and triplet levels.\footnote{We also reproduce the analysis in Figure~\ref{fig:scores_graph} after applying LGN in Appendix~\ref{appendix:results under the LGN strategy}.}
This directly addresses the concern raised in the system-level analysis: without normalization, averaging scores across directions is unreliable, as some languages may be systematically over- or under-estimated.
Our results suggest that applying LGN before aggregation provides a more reliable basis for multilingual system evaluation.
While the generalizability of LGN warrants further investigation, these findings offer initial evidence that normalization-based scoring can mitigate cross-lingual bias in automatic evaluation metrics.

\section{Conclusion}
In this work, we introduce XQ-MEval, the first multilingual dataset designed to achieve parallel quality across languages for benchmarking automatic evaluation metrics. 
Based on the benchmark, we identify limitations in the commonly used practice of averaging metric scores across translation directions to represent system-level performance.
Specifically, we reveal that cross-lingual scoring bias, caused by metrics exhibiting different scoring ranges across languages, is a key factor contributing to the misalignment between metrics and human evaluation in multilingual settings.
Building on this observation, we propose a normalization-based strategy to mitigate cross-lingual scoring bias by narrowing the distances between score ranges.
Experimental results show that the LGN strategy significantly improves the consistency with human evaluations and highlight the importance of aligning score ranges across languages to a unified scale before averaging for reliability.

\section*{Limitations}
Human evaluation remains a major bottleneck in machine translation research, as large-scale multilingual annotation, especially for expert-level annotation, is costly and resource-intensive.
Although our semi-automatic pipeline alleviates this reliance and makes benchmark construction more efficient, the current version covers only nine translation directions.
Nevertheless, the pipeline is highly flexible and can be extended to more languages in future work.

While the MQM framework provides a comprehensive set of error categories, we focus on only four purely semantic error types in our work.
However, as discussed in Section~\ref{subsuc:phrase construction}, these error types are better suited for achieving cross-lingual comparability and represent the most prominent categories in existing MQM datasets, accounting for approximately 46.3\% of all errors.
Although our pipeline can incorporate additional error types, doing so first requires careful linguistic justification to ensure that the added types remain comparable across languages.

Given that this is the first work to discuss the fairness in evaluating multilingual translation systems, our work raises further questions for future research.
For instance, are metrics equally sensitive to different error types, or do they respond unevenly?
More intriguingly, does this sensitivity vary across languages?
We leave these fine-grained investigations for future work.

\section*{Ethics Statement}
In this work, we construct the XQ-MEval dataset based on Flores, a public dataset, combining manual filtering to enhance its quality. 
We recruit eligible students from our institution to assist with human annotation tasks, and the compensation provided is in compliance with local standards.
All human-involved steps during the construction are carefully designed to ensure that no personal information is involved. 
The manual annotation process adheres strictly to the ethical guidelines of our institution and the ACL ethics policy.
Thus, this recruitment and annotation are approved by the ethics reviewing committee of our affiliation.
Generally, this benchmark can be applied in real-world scenarios, supporting the evaluation of automatic evaluation metrics in multilingual settings.

% Add license information following Flores 
Flores is released under the CC BY-SA 4.0 license\footnote{\url{https://huggingface.co/datasets/openlanguagedata/flores_plus}}, which explicitly permits adaptation and sharing.
To fully comply with these terms, our license in releasing XQ-MEval would be CC BY-SA 4.0.
% Add license information following OpenAI
Moreover, XQ-MEval is created using GPT-4o and is therefore subject to OpenAI's license terms\footnote{\url{https://openai.com/policies/terms-of-use}}.
OpenAI assigns to us all rights, titles, and interests in and to the output.

\section*{Use of AI Assistance}
During the preparation of this paper, we used ChatGPT to assist with proofreading and polishing.
The model was employed solely to improve clarity, grammar, and readability of the manuscript; all ideas, experimental designs, analyses, and conclusions come from the authors.
The authors carefully reviewed and verified all AI-assisted edits to ensure correctness and faithfulness to the intended meaning.
\bibliography{custom}
\appendix

\section{Computational Procedure}
\label{appendix:pseudocode}
Algorithm~\ref{alg:avg-strategy} formalizes the average strategy described in Section~\ref{sec:intro}, which evaluates multilingual MT systems by first computing metric scores for each triplet (Step~5) and then averaging scores across all translation directions to obtain a system-level score (Step~14).
Two highlighted components further clarify key aspects of our evaluation setup.
Step~15 computes the corresponding human score to serve as the predefined performance used to benchmark metrics against human judgments, as discussed in Section~\ref{subsec:data generation system-level}.
In addition, Step~7 present the normalization based LGN strategy proposed in Section~\ref{subsec:methodology}, where triplet level metric scores are normalized after computation.
\begin{algorithm}
\caption{Evaluation with Average Strategy}
\label{alg:avg-strategy}
\begin{algorithmic}[1]

\State \textbf{Input:}
number of language pairs $N$; number of triplets per language pair $I$;
metric scoring function $\textsc{Metric}(\tilde{r})$;
human scoring function
$\textsc{Human}(\tilde{r})$;
normalization flag \textsc{USE\_LGN};
normalization function $\textsc{LGN}(s_m)$

\State \textbf{Output:}
overall metric score $S_M$; overall human score $S_H$

\For{$i \leftarrow 1$ to $N$} \Comment{language pairs}
    \For{$j \leftarrow 1$ to $I$} \Comment{triplets}
        \State $s_m^{(j)} \leftarrow \textsc{Metric}(\tilde{r}_{i,j})$
        \If{\textsc{USE\_LGN}}
    \State \colorbox{red!15}{\strut $s_m^{(j)} \leftarrow \textsc{LGN}(s_m^{(j)})$}
\EndIf
        \State $s_h^{(j)} \leftarrow \textsc{Human}(\tilde{r}_{i,j})$
    \EndFor
    \State $\bar{s}_m^{(i)} \leftarrow \frac{1}{I} \sum_{j=1}^{I} s_m^{(j)}$
    \State $\bar{s}_h^{(i)} \leftarrow \frac{1}{I} \sum_{j=1}^{I} s_h^{(j)}$
\EndFor

\State $S_M \leftarrow \frac{1}{N} \sum_{i=1}^{N} \bar{s}_m^{(i)}$
\State \colorbox{yellow!45}{$S_H \leftarrow \frac{1}{N} \sum_{i=1}^{N} \bar{s}_h^{(i)}$}

\State \Return $S_M, S_H$

\end{algorithmic}
\end{algorithm}

\section{Language Selection in Benchmark Construction}
\label{appendix:language-selection}
In constructing the benchmark, we select nine target languages paired with English, resulting in nine translation directions: \texttt{en}-\texttt{zh}, \texttt{en}-\texttt{lo}, \texttt{en}-\texttt{ja}, \texttt{en}-\texttt{vi}, \texttt{en}-\texttt{id}, \texttt{en}-\texttt{es}, \texttt{en}-\texttt{fr}, \texttt{en}-\texttt{si}, and \texttt{en}-\texttt{de}.  
This selection aims to ensure a comprehensive evaluation across high-resource and low-resource languages.
As discussed in Section~\ref{section:related_work}, most widely-used metrics are driven by MQM-style training, i.e., fine-tuned on MQM-annotated data.  
However, MQM annotations are only available for high-resource languages, resulting in an imbalanced data distribution.  
Intuitively, this imbalance may lead MQM-driven metrics to exhibit stronger biases when evaluating translations in low-resource languages.
In addition, practical constraints such as the availability of native-speaking volunteers for filtering pseudo translations also influence our language choices.  
Taking these factors into account, we determine that the selected translation directions strike a reasonable balance between linguistic diversity and feasibility, making the benchmark both representative and manageable.
In addition, among the selected languages, those supported by MQM training data include: \texttt{zh}, \texttt{de}, \texttt{es}, \texttt{ja}, and \texttt{fr};
Languages without MQM support include: \texttt{lo}, \texttt{vi}, \texttt{id}, and \texttt{si}.

\section{Verification on MQM}
\label{appendix:preliminary experiment}
As mentioned in Section \ref{sec:intro}, investigating cross-lingual scoring bias requires instances with strictly parallel semantics and quality.
However, MQM datasets cover only a limited number of language pairs, among which only \texttt{en}-\texttt{de} and \texttt{en}-\texttt{ru} satisfy this requirement.
For these two directions, we partition instances into five MQM score ranges: $0$, $(0,5]$, $(5,10]$, $(10,15]$, and $(15,25]$, merging the highest range due to data sparsity. 
We evaluate these instances using BLEURT, XCOMET, and COMETKIWI-23 (spanning all metric types).

\begin{figure}[!ht]
    \centering
        \includegraphics[width=0.4\textwidth]{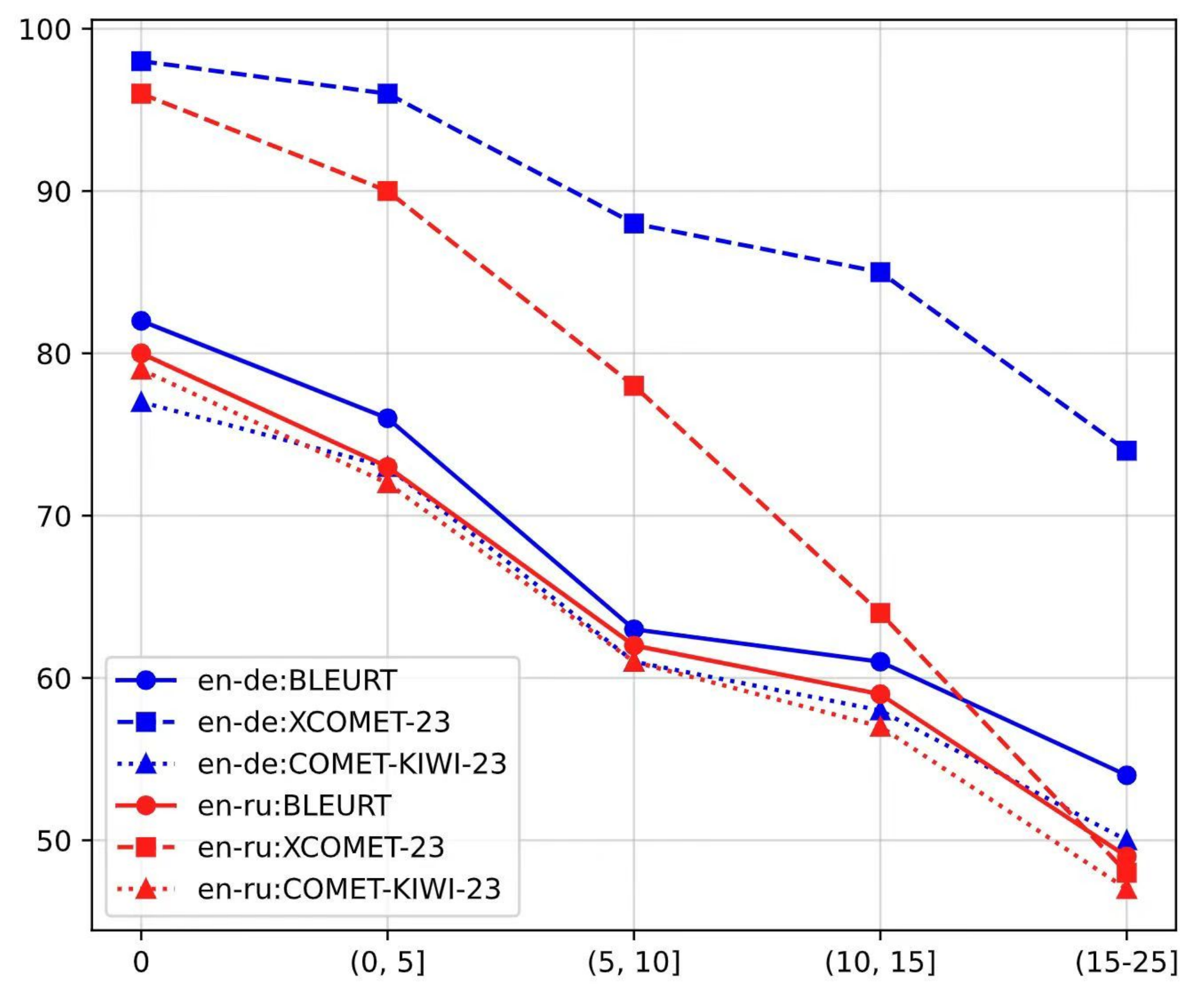}
    \caption{Visualization of three metrics scores across two directions at varying translation quality levels on MQM dataset.}
    \label{fig:preliminary experiment}
\end{figure}

The results in Figure~\ref{fig:preliminary experiment} show that translations of comparable quality in different language pairs are assigned different scores by the metrics, particularly XCOMET, even when only two language pairs are involved. 
Moreover, the results demonstrate that cross-lingual scoring bias exists in MQM data and follows a trend similar to that observed in Figure~\ref{fig:scores_graph}, thereby validating our synthetic instances.

In addition, we conduct additional experiments on real MQM datasets using the factors learned from our benchmark.
Specifically, we select \textit{mqm\_generalMT2024\_ende} and \textit{mqm\_generalMT2024\_enes} dataset, as these two language pairs overlap with those covered in our benchmark.
We observe a severe imbalance in the \texttt{en}-\texttt{es} data, where triplets with multiple errors are rare. 
To improve comparability, we restrict \texttt{en}-\texttt{de} to triplets with few errors.
The datasets contain outputs from different systems, which we rank using averaged MQM scores across language pairs. 
As no references are available, we use the reference-free metric COMETKIWI-23 for evaluation. 
We compute system-level rankings using both original and LGN-calibrated scores.
Calibration improves the correlation with MQM rankings \textbf{from 45.05 to 46.81}, indicating better alignment with human evaluation. 
Despite the limited setting, these results provide further evidence for the effectiveness of LGN, corroborating the validity of our synthetic instances.

\section{Annotation Guidelines}
\label{appendix:instruction}
To ensure that native speakers acquire a clear understanding of the purpose of our experiment and the definition of MQM, thereby enabling them to more accurately identify and filter error candidates that meet the required criteria, we comply an instruction document that provides the necessary background information and operational guidelines. 
It is included in the following:

\subsection*{Background}
In the evaluation of translation quality, a human-centric framework known as \textbf{Multidimensional Quality Metrics (MQM)} (\url{https://themqm.org/}) is widely used. Specifically, MQM classifies translation quality based on a standardized error taxonomy, resulting in a scoring system that is both low in subjectivity and high in comparability. This framework significantly facilitates both production and research efforts.

However, MQM annotation is inherently inefficient and costly, as it heavily depends on the manual work of expert annotators. While, in theory, advanced artificial intelligence could act as expert-level annotators, such a substitution is not entirely trustworthy because we cannot verify whether the AI has truly reached expert proficiency.

Fortunately, and interestingly, our task is \textbf{NOT} to evaluate a machine translation system in the MQM style. Instead, we aim to obtain MQM-style scores. Specifically, this means we can use advanced AI systems to disrupt a set of perfect translations by introducing errors defined under MQM. Then, we simply ask native speakers to verify whether the disruption was successful. This approach allows us to obtain reliable MQM scores on a given dataset.

\subsection*{Task}
Each volunteer will be provided with four files, named \texttt{en-\{lang\}-\{error\}.tsv}, where \texttt{\{lang\}} points to each volunteer’s native language, and \texttt{\{error\}} refers to four common and easily quantified types of errors in machine translation:
\textit{Addition}, \textit{Omission}, \textit{Mistranslation}, and \textit{Untranslated}.

In each file, there are three parts that should be noticed:
\begin{itemize}
    \item \textbf{src}: The source sentence in English.
    \item \textbf{ref}: The correct (perfect) translation of the source sentence in the volunteer’s native language.
    \item \textbf{mt}: The sentence that has been disrupted by using GPT-4o. Specifically, GPT-4o introduced an error into each \texttt{ref}.
\end{itemize}

Please note, the error in \texttt{mt} is marked by \texttt{<v> </v>}. Now, you should check the quality of \texttt{mt}, and judge whether the error marked by \texttt{<v> </v>} indeed disrupts \texttt{ref} without any change in the rest part of \texttt{ref}. If the answer is \textbf{YES}, you don't need to take any action; otherwise, you should write \texttt{T} in the reject column to indicate that the disruption is not acceptable.

\subsection*{Criteria}
The following are the evaluation criteria for each type of error:

\subsubsection*{Addition}
The error in \texttt{mt} marked by \texttt{<v></v>} introduced additional semantics into \texttt{ref}.
\begin{itemize}
    \item If the error indeed presents additional semantics in the \texttt{ref} without any change in the rest part of \texttt{ref}, then this \texttt{mt} is acceptable, i.e., you don't need to take any action.
    \item Otherwise, please write \texttt{T} in the reject column.
    \item Note that the key is whether the semantics are changed, i.e., a change in the adverb of degree is considered as reject.
\end{itemize}

\subsubsection*{Omission}
The \texttt{mt} has a missing part compared to the \texttt{ref}, and the missing part is marked by \texttt{<v></v>}.
\begin{itemize}
    \item If the missing part in the \texttt{mt} causes a change in meaning, then this \texttt{mt} is acceptable, i.e., you don't need to take any action.
    \item Note, omission could make the sentence unreadable. However, the unique criterion is that the part outside of labeling (\texttt{<v></v>}) is not changed.
    \item In languages using spaces as intervals, some words could be labeled. However, the following case caused by changes of punctuation is also acceptable:
    \begin{quote}
        \texttt{ref}: ..., en particulier les affaires de voitures volées, avec l'intention...\\
        \texttt{mt}: ..., en <v>particulier,</v> avec l'intention...
    \end{quote}
    Here, \texttt{les affaires de voitures volées} is omissive, and the label is caused by the change of \texttt{particulier} $\rightarrow$ \texttt{particulier,}.
    \item Otherwise, e.g., the missing part changes the part out of \texttt{<v></v>} or the marked part is not missed, please write \texttt{T} in the reject column.
\end{itemize}

\subsubsection*{Mistranslation}
The error in \texttt{mt} marked by \texttt{<v></v>} is a mistranslation from \texttt{src}.
\begin{itemize}
    \item Given that \texttt{ref} is a ground-truth translation from \texttt{src}, you can simply compare \texttt{ref} and \texttt{mt}. If the error of \texttt{mt} conveys different words or semantics compared to \texttt{ref}, this \texttt{mt} is acceptable, i.e., you don't need to take any action.
    \item Otherwise, please write \texttt{T} in the reject column.
\end{itemize}

\subsubsection*{Untranslated}
The error in \texttt{mt} marked by \texttt{<v></v>} has not been translated and remains in the original English.
\begin{itemize}
    \item Simply copying from \texttt{src} or changing words but remaining in English is recognized as acceptable, i.e., you don't need to take any action.
    \item If the untranslated words are person’s names or place names, please write \texttt{T} in the reject column.
\end{itemize}

\subsection*{Overall}
Changes in the content of the \texttt{mt} may result in grammatical errors in the overall sentence, and this is acceptable as long as the part marked with \texttt{<v></v>} in the \texttt{mt} indeed causes a change in meaning without changes in the part outside of \texttt{<v></v>}. This indicates that the \texttt{mt} is acceptable.

\section{Prompt Design}
\label{appendix:prompt}
\begin{figure*}[!ht]
    \centering
        \includegraphics[width=1.0\textwidth]{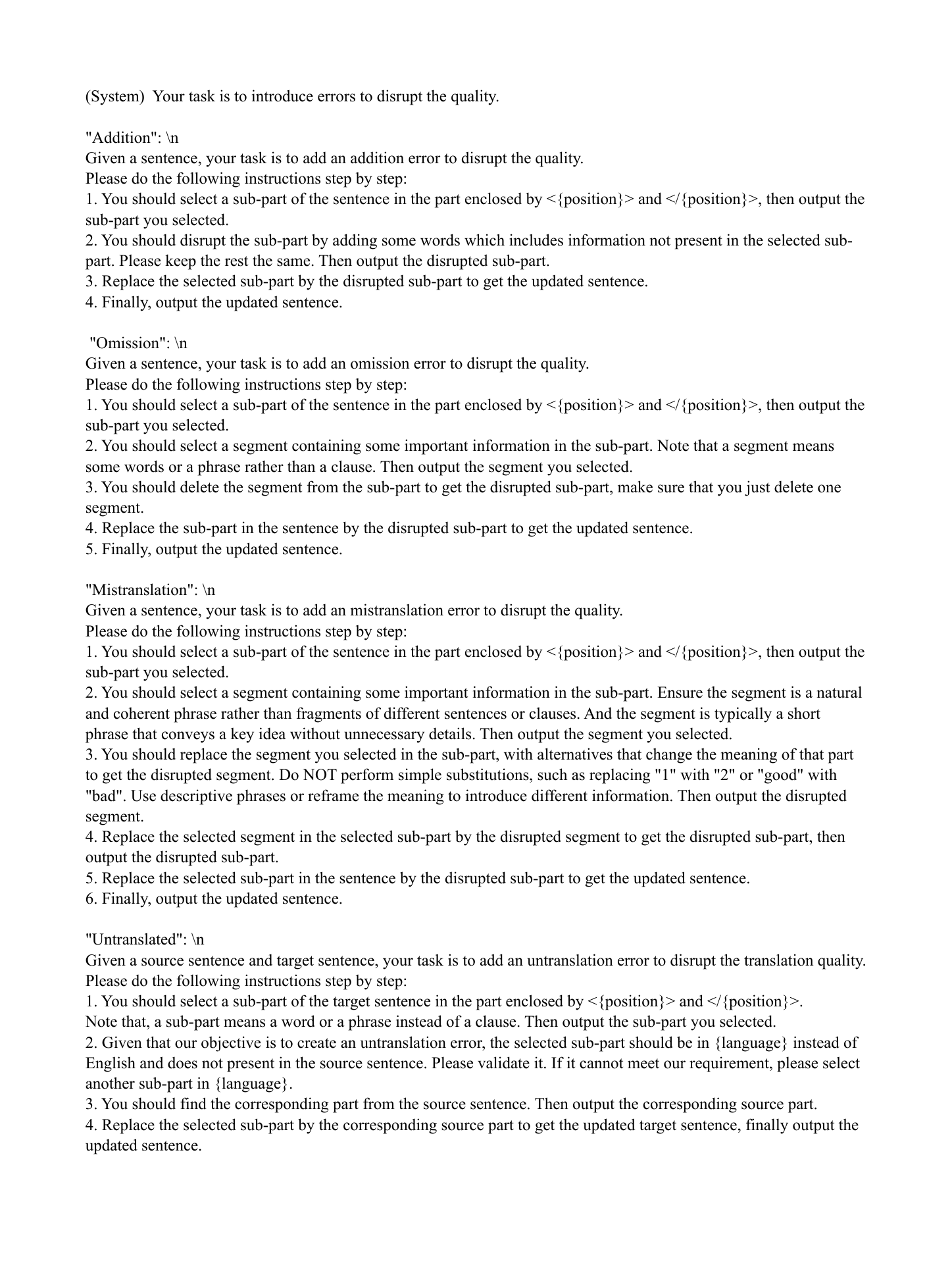}
    \vspace{-5.3em}
    \caption{The prompt for different error types to guide GPT-4o to introduce errors to references.}
    \label{fig:prompt}
\end{figure*}

To instruct GPT-4o to introduce addition, omission, mistranslation and untranslated errors to references to obtain temporary error candidates containing one error segment, we design the specific prompt for different error types. 
Figure~\ref{fig:prompt} shows the details of the prompt.

\section{Triplets Count Distribution}
\label{appendix:numbers of sentences of 5 levels}
Table~\ref{tab:numbers of all combinations} shows the triplets count distribution across the five quality levels for each language pair.
As shown in the table, the triplets with quality level 2 and 3 are more frequent, while triplets at level 5 are fewer. 
This is because quality level reflects the number of errors in pseudo translations; as the error count increases, overlapping error spans reduce the number of generated triplets.
\begin{table}[t]
    \centering
    \resizebox{\linewidth}{!}{
    \begin{tabular}{crrrrrrrrr}
    \hline
        
        Quality Level & \texttt{en}-\texttt{zh} & \texttt{en}-\texttt{lo} & \texttt{en}-\texttt{ja} & \texttt{en}-\texttt{vi} & \texttt{en}-\texttt{id} & \texttt{en}-\texttt{fr} & \texttt{en}-\texttt{es} & \texttt{en}-\texttt{si} & \texttt{en}-\texttt{de}  \\ \hline
        1 & 776 & 753 & 775 & 771 & 782 & 775 & 771 & 765 & 774  \\ 
        2 & 2,109 & 2,053 & 2,078 & 2,056 & 2,095 & 1,992 & 2,016 & 2,064 & 2,049  \\ 
        3 & 2,548 & 2,627 & 2,441 & 2,420 & 2,421 & 2,068 & 2,233 & 2,489 & 2,337  \\ 
        4 & 1,466 & 1,704 & 1,324 & 1,387 & 1,311 & 9,57 & 1,069 & 1,432 & 1,234 \\ 
        5 & 406 & 558 & 340 & 428 & 312 & 198 & 203 & 361 & 313  \\ \hline
    \end{tabular}
    }
    \caption{The triplets count distribution across the five quality levels for each language pair.}
    \label{tab:numbers of all combinations}
\end{table}
\begin{table}[t]
    \centering
    \begin{tabular}{crr}
    \hline
        Num. & System-level &Triplet-level  \\ \hline
        3 & 0.03 & 0.03  \\ 
        6 & 0.009 & 0.003  \\ 
        9 & 0.001 & 0.005 \\ \hline
    \end{tabular}
    \caption{Paired samples t-test results for system-level and triplet-level improvements obtained with the LGN strategy.}
    \label{tab:significance test}
\end{table}

\section{Discussion on Repeated Sampling}
\label{appendix: mean and variance for 5, 10, 25 sampling}
To examine the effect of repeated sampling on evaluation stability, we test three metrics, i.e., BLEURT, xCOMET, and KIWI23, on monolingual systems for \texttt{en}-\texttt{zh}, \texttt{en}-\texttt{ja}, and \texttt{en}-\texttt{de} at five quality levels.
For each system, 102 triplets are sampled, and the procedure is repeated 5, 10, and 25 times.
Table~\ref{tab:main_table for mean and var} reports the means and variances across these settings.
As the sampling iterations increase, the mean scores shown in the table exhibit stability.
Although the variance fluctuates to some extent, it is caused by the value is scaled to the square of the scoring scale because the scores are amplified by a factor of 100. 
Consequently, the variance remains within a small range, and we consider these fluctuations to be acceptable.
Ultimately, we adopt the approach of repeating the process 10 times in our main experiments.

\begin{table*}[t]
    \centering
    \begin{subtable}[t]{1.0\linewidth}
        \centering
        \resizebox{\textwidth}{!}{
        \begin{tabular}{cccccccccccccccccc}
            \hline
            \textbf{} & \textbf{} & \textbf{} & \textbf{BLEURT}  & \textbf{} & \textbf{} & \textbf{} & \textbf{} & \textbf{} & \textbf{xCOMET} & \textbf{} & \textbf{} & \textbf{} & \textbf{} & \textbf{} & \textbf{KIWI23} & \textbf{ } & \textbf{} \\ \hline
            \textbf{mean} & 1 & 2  & 3  & 4  & 5 & ~ & 1  & 2  & 3  & 4  & 5 & ~ & 1  & 2  & 3  & 4  & 5  \\ \hdashline
            \texttt{en}-\texttt{zh} & 81.98  & 70.37  & 62.36  & 56.45  & 51.85 & ~ & 82.90  & 68.44  & 55.50  & 45.20  & 36.66 & ~ & 66.38  & 56.09  & 46.43  & 40.23  & 33.39   \\ 
            \texttt{en}-\texttt{ja} & 81.65  & 69.15  & 60.18  & 53.47  & 48.18 & ~ & 80.85  & 63.93  & 49.20  & 38.78  & 32.51 & ~ & 67.15  & 57.28  & 49.18  & 43.50  & 39.67    \\ 
            \texttt{en}-\texttt{de} & 80.05  & 67.82  & 59.01  & 52.02  & 47.85 & ~ & 91.89  & 83.21  & 72.83  & 64.83  & 56.65 & ~ & 61.92  & 49.16  & 38.19  & 29.52  & 25.43   \\ 
            \hline
            \textbf{var} & 1  & 2  & 3  & 4  & 5 & ~ & 1 & 2  & 3  & 4  & 5 & ~ & 1  & 2  & 3  & 4  & 5   \\ \hdashline
            \texttt{en}-\texttt{zh} & 0.36  & 0.36  & 1.49  & 0.40  & 0.62 & ~ & 0.55  & 0.85  & 1.27  & 1.21  & 0.74 & ~ & 0.33  & 0.69  & 1.04  & 0.07  & 1.68    \\ 
            \texttt{en}-\texttt{ja} & 0.41  & 0.58  & 0.10  & 0.80  & 0.83 & ~ & 0.35  & 2.22  & 2.99  & 3.39  & 1.60 & ~ & 0.27  & 1.44  & 0.89  & 0.40  & 1.43   \\ 
            \texttt{en}-\texttt{de} & 0.39  & 1.94  & 1.22  & 1.83  & 0.37 & ~ & 0.19  & 0.26  & 2.71  & 2.57  & 4.01 & ~ & 0.94  & 1.38  & 3.70  & 1.43  & 1.68   \\ \hline
        \end{tabular}
        }
        \caption{Mean and variance for 5 iterations of sampling.}
        \label{tab:subtable_5}
    \end{subtable}
    \vspace{1.0em}
    \begin{subtable}[t]{1.0\linewidth}
        \centering
        \resizebox{\textwidth}{!}{
        \begin{tabular}{cccccccccccccccccc}
            \hline
            \textbf{} & \textbf{} & \textbf{} & \textbf{BLEURT} & \textbf{} & \textbf{} & \textbf{} & \textbf{} & \textbf{} & \textbf{xCOMET} & \textbf{} & \textbf{} & \textbf{} & \textbf{} & \textbf{} & \textbf{KIWI23} & \textbf{} & \textbf{}  \\ \hline
            \textbf{mean} & 1 & 2 & 3 & 4 & 5 & ~ & 1 & 2 & 3 & 4 & 5 & ~ & 1 & 2 & 3 & 4 & 5   \\ \hdashline
            \texttt{en}-\texttt{zh} & 81.90  & 70.60  & 62.42  & 56.36  & 52.09 & ~ & 82.93  & 68.49  & 56.06  & 45.32  & 37.41 & ~ & 66.28  & 56.12  & 46.95  & 39.87  & 33.83   \\ 
            \texttt{en}-\texttt{ja} & 81.97  & 69.27  & 60.36  & 53.74  & 48.21 & ~ & 81.12  & 64.38  & 48.79  & 38.62  & 32.48 & ~ & 67.66  & 57.38  & 49.36  & 43.73  & 39.54    \\ 
            \texttt{en}-\texttt{de} & 79.63  & 67.99  & 59.62  & 52.23  & 47.35 & ~ & 91.45  & 83.20  & 73.64  & 64.12  & 55.40 & ~ & 60.92  & 48.82  & 38.95  & 29.87  & 24.84  \\ 
            \hline
            \textbf{var} & 1  & 2  & 3  & 4  & 5 & ~ & 1  & 2  & 3  & 4  & 5 & ~ & 1  & 2  & 3  & 4  & 5    \\ \hdashline
            \texttt{en}-\texttt{zh} & 0.22  & 0.42  & 1.00  & 0.36  & 0.62 & ~ & 0.67  & 0.60  & 2.36  & 0.85  & 1.02 & ~ & 0.29  & 0.41  & 2.05  & 0.17  & 1.66    \\ 
            \texttt{en}-\texttt{ja} & 0.45  & 0.43  & 0.36  & 0.89  & 0.58 & ~ & 0.79  & 1.56  & 2.26  & 2.87  & 1.28 & ~ & 0.72  & 0.76  & 0.69  & 0.59  & 0.94     \\ 
            \texttt{en}-\texttt{de} & 0.71  & 1.33  & 1.52  & 1.18  & 0.64 & ~ & 0.80  & 0.57  & 4.47  & 2.24  & 6.12 & ~ & 1.84  & 1.54  & 4.35  & 0.99  & 1.95   \\ \hline
        \end{tabular}
        }
        \caption{Mean and variance for 10 iterations of sampling.}
        \label{tab:subtable_10}
    \end{subtable}
    \vspace{1.0em}
    \begin{subtable}[t]{1.0\linewidth}
    \centering
    \resizebox{\textwidth}{!}{
        \begin{tabular}{cccccccccccccccccc}
        \hline
        \textbf{} & \textbf{} & \textbf{} & \textbf{BLEURT} & \textbf{} & \textbf{} & \textbf{} & \textbf{} & \textbf{} & \textbf{xCOMET} & \textbf{} & \textbf{} & \textbf{} & \textbf{} & \textbf{} & \textbf{KIWI23} & \textbf{} & \textbf{ }  \\ \hline
        \textbf{mean} & 1 & 2 & 3 & 4 & 5 & ~ & 1 & 2 & 3 & 4 & 5 & ~ & 1 & 2 & 3 & 4 & 5   \\ \hdashline
        \texttt{en}-\texttt{zh} & 81.80  & 70.52  & 62.39  & 56.21  & 51.91  & ~ & 82.56  & 68.64  & 55.88  & 45.51  & 37.23  & ~ & 66.10  & 55.66  & 46.72  & 39.93  & 33.59   \\ 
        \texttt{en}-\texttt{ja} & 81.98  & 69.71  & 60.58  & 53.65  & 47.92  & ~ & 81.34  & 64.51  & 49.80  & 38.56  & 32.04  & ~ & 67.72  & 57.55  & 49.73  & 43.68  & 39.23   \\ 
        \texttt{en}-\texttt{de} & 79.63  & 68.04  & 59.47  & 52.91  & 47.29  & ~ & 91.41  & 83.23  & 73.69  & 64.79  & 55.16  & ~ & 60.89  & 48.54  & 38.51  & 30.84  & 24.78    \\ 
        \hline
        \textbf{var} & 1 & 2 & 3 & 4 & 5 & ~ & 1 & 2 & 3 & 4 & 5 & ~ & 1 & 2 & 3 & 4 & 5   \\ \hdashline
        \texttt{en}-\texttt{zh} & 0.34  & 0.53  & 1.07  & 0.64  & 0.78  & ~ & 0.77  & 1.52  & 2.67  & 1.37  & 0.88  & ~ & 0.43  & 1.49  & 1.82  & 1.01  & 1.10    \\ 
        \texttt{en}-\texttt{ja} & 0.42  & 0.49  & 0.81  & 0.73  & 0.71  & ~ & 1.87  & 1.86  & 3.87  & 2.37  & 1.77  & ~ & 0.90  & 0.56  & 1.89  & 0.79  & 1.25  \\ 
        \texttt{en}-\texttt{de} & 0.50  & 0.99  & 1.25  & 1.49  & 0.78  & ~ & 0.59  & 1.01  & 2.48  & 3.04  & 3.75  & ~ & 1.49  & 1.92  & 3.15  & 2.95  & 1.31 \\ \hline
        \end{tabular}
        }
        \caption{Mean and variance for 25 iterations of sampling.}
        \label{tab:subtable_25}
   \end{subtable}
   \caption{Mean and variance for 5, 10, 25 iterations of sampling. Note that the scores are amplified by a factor of 100, and the scale of the variance corresponds to the square of the scoring scale.}
   \label{tab:main_table for mean and var}
\end{table*}

\section{Detailed Scores}
\label{appendix:graphs and detailed scores}

Table~\ref{tab:new detailed scores} presents the detailed scores of Figure \ref{fig:scores_graph}.
\begin{table*}[!ht]
    \centering
    \begin{subtable}[t]{1.0\linewidth}
        \centering
        \resizebox{\textwidth}{!}{
        \begin{tabular}{cccccccccccccccccccccccccccccccccccccccccccccccccccccc}
            \hline
        \textbf{} & \textbf{} & \textbf{} & \textbf{spBLEU} & \textbf{} & \textbf{} & \textbf{} & \textbf{} & \textbf{} & \textbf{chrF} & \textbf{} & \textbf{} & \textbf{} & \textbf{} & \textbf{} & \textbf{BLEURT} & \textbf{} & \textbf{}  \\ \hline
        \textbf{} & 1 & 2 & 3 & 4 & 5 & ~ & 1 & 2 & 3 & 4 & 5 & ~ & 1 & 2 & 3 & 4 & 5\\ \hline
        \texttt{en}-\texttt{zh} & 83.85  & 70.50  & 60.36  & 51.84  & 45.16  & ~ & 74.46  & 62.94  & 54.34  & 46.97  & 41.70  & ~ & 81.90  & 70.81  & 62.60  & 56.89  & 52.61  \\
        \texttt{en}-\texttt{lo} & 84.99  & 73.74  & 64.88  & 55.64  & 48.21  & ~ & 87.33  & 77.31  & 69.93  & 63.03  & 57.35  & ~ & 82.87  & 73.03  & 65.26  & 57.72  & 52.88 \\
        \texttt{en}-\texttt{ja} & 81.13  & 66.94  & 55.20  & 45.00  & 35.78  & ~ & 75.03  & 65.13  & 57.05  & 50.47  & 44.38  & ~ & 81.97  & 69.74  & 61.05  & 54.29  & 47.59 \\
        \texttt{en}-\texttt{vi} & 84.21  & 72.71  & 62.93  & 56.48  & 51.55  & ~ & 90.44  & 82.54  & 75.98  & 71.39  & 67.88  & ~ & 81.85  & 71.86  & 63.54  & 57.29  & 52.25\\
        \texttt{en}-\texttt{id} & 83.10  & 70.02  & 60.12  & 51.25  & 44.98  & ~ & 90.83  & 82.89  & 76.89  & 71.49  & 67.46  & ~ & 83.15  & 74.56  & 68.42  & 64.55  & 62.11\\
        \texttt{en}-\texttt{fr} & 84.03  & 71.60  & 62.46  & 55.21  & 48.60  & ~ & 90.41  & 82.03  & 75.74  & 70.37  & 65.28  & ~ & 80.96  & 67.37  & 56.23  & 47.83  & 40.07 \\
        \texttt{en}-\texttt{es} & 84.30  & 71.72  & 62.35  & 53.94  & 46.73  & ~ & 90.80  & 82.63  & 76.49  & 71.52  & 67.16  & ~ & 83.37  & 71.68  & 62.18  & 54.59  & 48.96 \\
        \texttt{en}-\texttt{si} & 86.18  & 75.42  & 67.07  & 59.80  & 53.27  & ~ & 91.40  & 83.41  & 77.29  & 71.65  & 66.73  & ~ & 84.38  & 78.44  & 72.83  & 68.57  & 64.73 \\
        \texttt{en}-\texttt{de} & 84.09  & 72.25  & 62.61  & 54.31  & 48.01  & ~ & 90.93  & 83.13  & 76.58  & 71.38  & 66.47  & ~ & 79.63  & 68.03  & 59.56  & 53.64  & 46.88 \\ \hline
        \end{tabular}
        }
        \caption{Sequenced-based metrics.}
        \label{tab:subtable_sequenced}
    \end{subtable}
    \vspace{1.0em}
    \begin{subtable}[t]{1.0\linewidth}
        \centering
        \resizebox{\textwidth}{!}{
        \begin{tabular}{cccccccccccccccccccccccccccccccccccccccccccccccccccccc}
            \hline
        \textbf{} & \textbf{} & \textbf{} & \textbf{COMET} & \textbf{} & \textbf{} & \textbf{} & \textbf{} & \textbf{} & \textbf{xCOMET} & \textbf{} & \textbf{} & \textbf{} & \textbf{} & \textbf{} & \textbf{MX-reg} & \textbf{} & \textbf{}  \\ \hline
        \textbf{} & 1 & 2 & 3 & 4 & 5 & ~ & 1 & 2 & 3 & 4 & 5 & ~ & 1 & 2 & 3 & 4 & 5\\ \hline
        \texttt{en}-\texttt{zh} & 90.42  & 85.97  & 82.17  & 78.46  & 75.10  & ~ & 82.93  & 68.65  & 55.82  & 45.86  & 37.67  & ~ & 2.40  & 3.58  & 4.75  & 5.63  & 6.86 \\
        \texttt{en}-\texttt{lo} & 88.15  & 83.31  & 79.07  & 74.60  & 71.16  & ~ & 63.72  & 48.64  & 36.11  & 27.04  & 21.01  & ~ & 3.46  & 5.49  & 7.24  & 8.81  & 10.22 \\
        \texttt{en}-\texttt{ja} & 92.63  & 88.90  & 85.18  & 81.85  & 77.12  & ~ & 81.12  & 64.52  & 49.83  & 39.70  & 31.86  & ~ & 1.85  & 2.91  & 3.87  & 4.91  & 6.03 \\
        \texttt{en}-\texttt{vi} & 89.81  & 85.61  & 81.59  & 78.68  & 75.26  & ~ & 81.83  & 68.44  & 55.56  & 46.71  & 36.54  & ~ & 3.22  & 5.05  & 6.55  & 8.05  & 9.71 \\
        \texttt{en}-\texttt{id} & 91.72  & 87.93  & 83.95  & 80.52  & 77.35  & ~ & 84.43  & 70.58  & 55.69  & 45.30  & 36.19  & ~ & 2.52  & 3.89  & 5.53  & 6.65  & 7.58 \\
        \texttt{en}-\texttt{fr} & 86.49  & 80.36  & 74.45  & 69.70  & 64.79  & ~ & 82.18  & 66.13  & 51.27  & 40.37  & 30.00  & ~ & 2.64  & 4.04  & 5.51  & 6.81  & 8.40 \\
        \texttt{en}-\texttt{es} & 87.15  & 81.37  & 75.73  & 70.81  & 66.54  & ~ & 86.06  & 72.62  & 59.45  & 47.80  & 37.82  & ~ & 2.44  & 3.90  & 5.40  & 6.37  & 7.28 \\
        \texttt{en}-\texttt{si} & 92.51  & 89.30  & 85.92  & 82.84  & 79.64  & ~ & 69.67  & 52.42  & 37.07  & 26.85  & 20.31  & ~ & 2.30  & 4.08  & 6.08  & 7.67  & 9.83 \\
        \texttt{en}-\texttt{de} & 87.39  & 81.18  & 75.75  & 70.89  & 64.36  & ~ & 91.45  & 83.13  & 73.95  & 66.14  & 55.02  & ~ & 1.53  & 2.23  & 2.87  & 3.40  & 4.17 \\ \hline
        \end{tabular}
        }
        \caption{Regression-based metrics.}
        \label{tab:subtable_regression}
    \end{subtable}
    \vspace{1.0em}
    \begin{subtable}[t]{1.0\linewidth}
        \centering
        \resizebox{\textwidth}{!}{
        \begin{tabular}{cccccccccccccccccccccccccccccccccccccccccccccccccccccc}
            \hline
        \textbf{} & \textbf{} & \textbf{} & \textbf{KIWI22} & \textbf{} & \textbf{} & \textbf{} & \textbf{} & \textbf{} & \textbf{KIWI23} & \textbf{} & \textbf{} & \textbf{} & \textbf{} & \textbf{} & \textbf{MX-qe} & \textbf{} & \textbf{}  \\ \hline
        \textbf{} & 1 & 2 & 3 & 4 & 5 & ~ & 1 & 2 & 3 & 4 & 5 & ~ & 1 & 2 & 3 & 4 & 5\\ \hline
        \texttt{en}-\texttt{zh} & 76.40  & 69.10  & 63.34  & 58.68  & 54.74  & ~ & 66.28  & 55.69  & 46.63  & 40.37  & 34.33  & ~ & 1.70  & 2.32  & 3.00  & 3.54  & 4.26   \\ 
        \texttt{en}-\texttt{lo} & 74.50  & 68.24  & 63.82  & 60.12  & 57.88  & ~ & 61.41  & 52.93  & 46.59  & 40.60  & 36.18  & ~ & 2.71  & 3.87  & 4.83  & 5.81  & 6.62  \\ 
        \texttt{en}-\texttt{ja} & 79.37  & 72.80  & 67.69  & 63.90  & 60.15  & ~ & 67.66  & 57.27  & 49.42  & 44.60  & 38.96  & ~ & 1.38  & 2.04  & 2.65  & 3.25  & 4.19  \\ 
        \texttt{en}-\texttt{vi}  & 76.38  & 70.37  & 65.30  & 61.81  & 58.60  & ~ & 64.94  & 55.71  & 48.44  & 43.27  & 38.43  & ~ & 2.04  & 2.89  & 3.62  & 4.43  & 5.11   \\ 
        \texttt{en}-\texttt{id} & 76.09  & 69.15  & 63.20  & 58.60  & 55.54  & ~ & 66.37  & 56.34  & 47.66  & 41.96  & 37.25  & ~ & 2.23  & 3.12  & 4.16  & 4.94  & 5.54  \\ 
        \texttt{en}-\texttt{fr} & 78.60  & 70.98  & 64.87  & 60.42  & 55.42  & ~ & 60.82  & 47.49  & 38.08  & 30.50  & 24.20  & ~ & 1.97  & 2.83  & 3.85  & 4.70  & 5.94  \\ 
        \texttt{en}-\texttt{es} & 76.16  & 68.48  & 61.98  & 57.04  & 53.43  & ~ & 64.96  & 53.30  & 43.91  & 36.34  & 30.66  & ~ & 1.97  & 2.84  & 3.87  & 4.63  & 5.26   \\ 
        \texttt{en}-\texttt{si} & 80.42  & 72.71  & 65.64  & 60.72  & 56.20  & ~ & 74.13  & 64.18  & 55.82  & 49.46  & 44.87  & ~ & 1.25  & 1.89  & 2.74  & 3.72  & 4.93   \\ 
        \texttt{en}-\texttt{de} & 75.62  & 68.56  & 62.28  & 58.17  & 53.91  & ~ & 60.92  & 48.52  & 38.47  & 31.73  & 24.32  & ~ & 1.44  & 2.11  & 2.71  & 3.35  & 4.21   \\ \hline    
        \end{tabular}
        }
        \caption{Regression-free metrics.}
        \label{tab:subtable_reference}
    \end{subtable}
    \caption{The detailed scores of nine metrics when evaluating different languages at various quality levels.}
   \label{tab:new detailed scores}
   
\end{table*}

\section{Score Reduction across Directions and Metrics}
\label{appendix:detailed explanation for score decline}

Figure~\ref{fig:scores_graph} reveals that as translation quality declines, the rate of score reduction differs across translation directions, highlighting the varying sensitivities of metrics to quality changes across languages. 
This variation exacerbates score inconsistencies across directions at the same quality level, particularly for lower-quality translations. 
The widening gaps in the decline patterns further illustrate this trend.
Similarly, score reduction patterns differ across metrics.
For spBLEU, scores are approaching at high quality but diverge as quality decreases due to different decline rates across directions.
chrF shows more consistent decline trends, though its score ranges vary substantially across directions, with \texttt{zh} and \texttt{ja} exhibiting systematically lower ranges.
BLEURT exhibits behavior similar to spBLEU, but with larger cross-lingual discrepancies in score reduction as translation quality deteriorates.
For COMET and xCOMET, score reduction trends exhibit similar patterns across directions. 
However, COMET assigns direction-specific score ranges with limited overlap, whereas xCOMET produces more aligned score ranges for most directions, except \texttt{lo}, \texttt{si}, and \texttt{de}.
In contrast, KIWI22 and KIWI23 more closely align with the desired properties of an ideal metric, as they exhibit more closely aligned score ranges and score reduction trends, whereas KIWI23 still shows noticeable score range discrepancies for certain directions.
By comparison, the MetricX variants display substantial cross-lingual inconsistency in both score ranges and reduction patterns, with regression-based MetricX exhibiting pronounced inconsistencies.

\section{Experiment on LLM-based Metrics}
\label{appendix:llm-as-judge}
\begin{figure*}[!ht]
    \centering
        \includegraphics[width=0.9\linewidth]{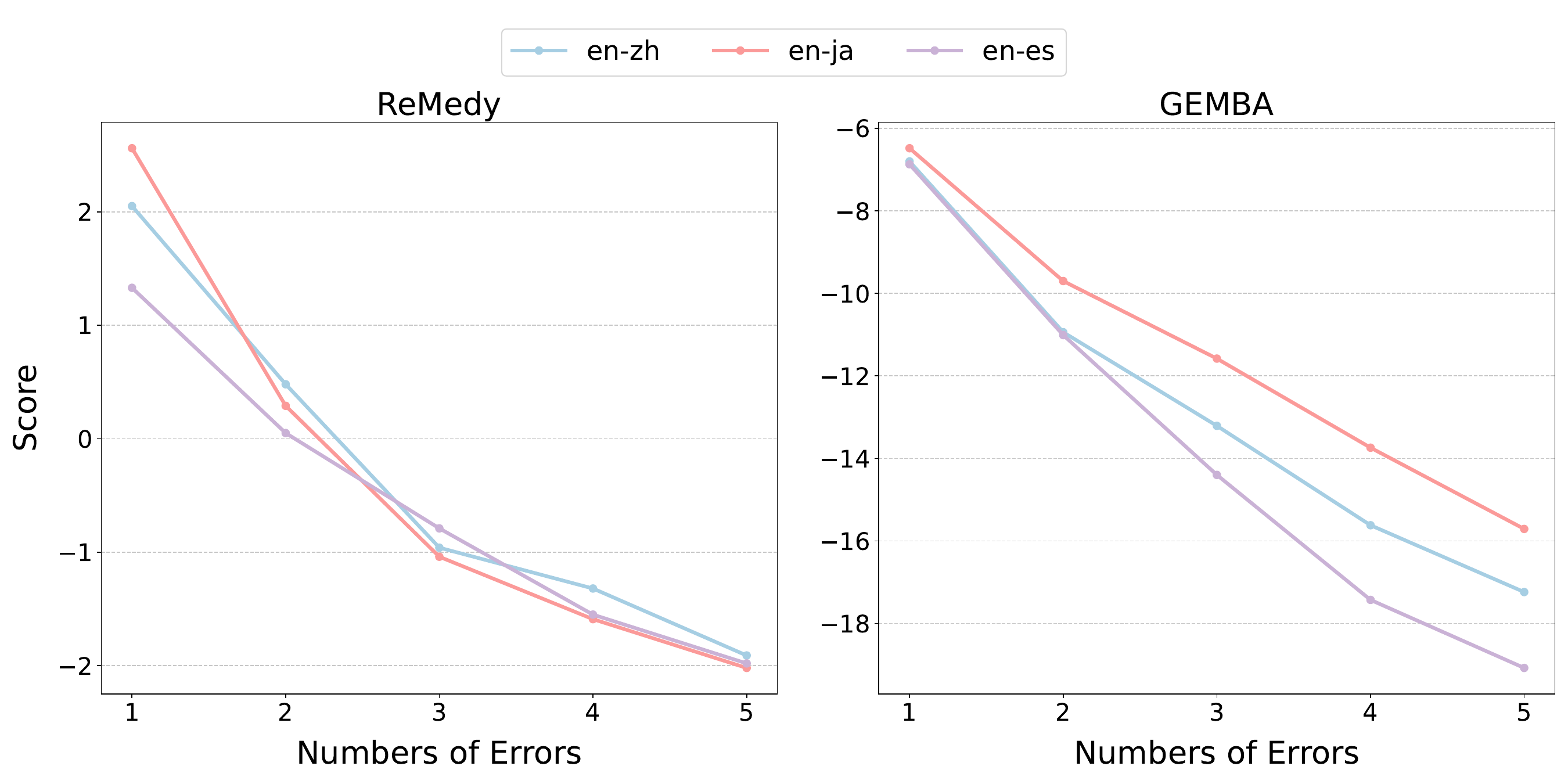}
    \vspace{-1em}
    \caption{
    Visualization of LLM-based evaluation scores across three directions at varying translation quality levels. 
}
    \vspace{-1.5em}
    \label{fig:llm metric}
\end{figure*}

We investigate two LLM-based evaluation approaches: ReMedy \cite{tan-monz-2025-remedy}, a trainable evaluation metric fine-tuned from LLMs; and GEMBA-MQM \cite{kocmi-federmann-2023-gemba}, which prompts LLMs to simulate human annotators by following MQM guidelines. 
Using these evaluators, we assess translation triplets at varying quality levels across three language pairs, \texttt{en}-\texttt{zh}, \texttt{en}-\texttt{es}, and \texttt{en}-\texttt{ja}, reflecting the language coverage of ReMedy in our study. 
The results in Figure~\ref{fig:llm metric} reveal substantial variation across language pairs, indicating that LLM-based evaluators remain susceptible to cross-lingual bias.

\section{Significance Test}
\label{appendix:significance test}
We conduct paired samples t-tests on the improvements obtained with the LGN strategy in Table~\ref{tab:normalized correlation}. As shown in the Table~\ref{tab:significance test}, all p-values are below 0.05, indicating that although the improvements are small in magnitude, they are statistically significant.

\section{Results under the LGN Strategy}
\label{appendix:results under the LGN strategy}
\begin{figure*}[!ht]
    \centering
        \includegraphics[scale=0.28]{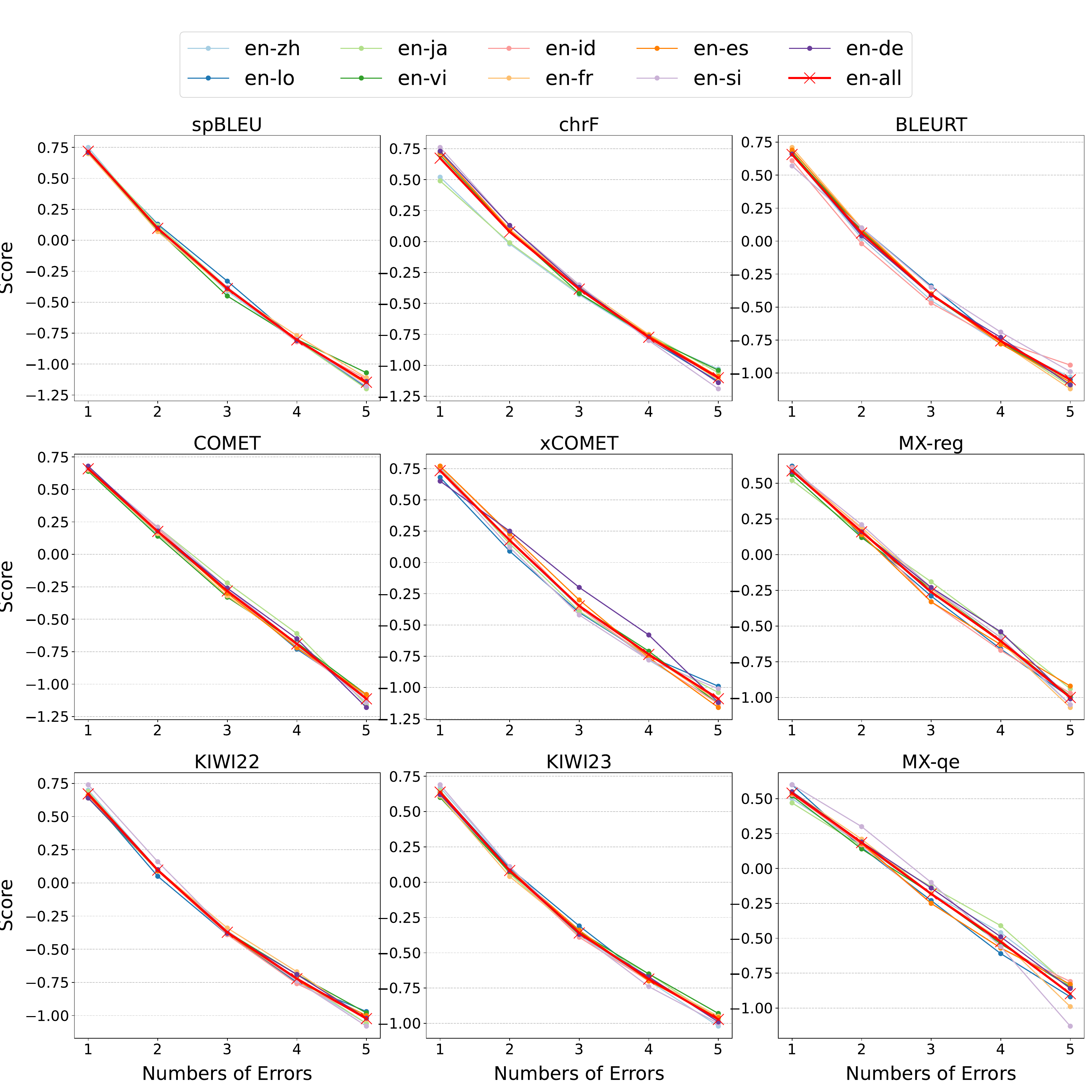}
    \vspace{-0.5em}
    \caption{Visualization of nine metrics scores under the LGN strategy across nine directions at varying translation quality levels.
    \texttt{en}-\texttt{all} denoting the average metric scores among all directions.}
    \vspace{-1em}
    \label{fig:n_scores_graph}
\end{figure*}

Figure~\ref{fig:n_scores_graph} shows the normalized scores of nine metrics across translation directions at varying quality levels. 
As illustrated in the figure, the LGN strategy effectively narrows score range disparities across language pairs, as evidenced by the reduced distances between curves.
After applying LGN, translations of comparable quality from different language pairs receive consistent metric scores, and the score degradation trends as translation quality decreases become more consistent across directions.

\end{document}